\documentclass[runningheads]{llncs}
\usepackage{graphicx}
\usepackage[letterpaper]{geometry}
\usepackage{amssymb}
\usepackage[T1]{fontenc}
\usepackage[utf8]{inputenc}
\usepackage{amsmath,mathtools}

\usepackage{amsthm}
\usepackage{color}
\usepackage{xcolor}
\usepackage{soul}
\usepackage{algpseudocode}
\usepackage{algorithm}
\usepackage{cite}
\usepackage{url}
\usepackage{hyperref}
\usepackage{enumitem}
\usepackage{makecell}
\usepackage{multicol}
\usepackage{subfigure}
\usepackage{wrapfig}
\usepackage{lipsum,booktabs}

\usepackage[final]{changes}

\DeclareMathOperator\supp{supp}
\newcommand{\reals}{\mathbb{R}}
\newcommand{\vc}[1]{\mathbf{#1}}
\newcommand{\cvx}{\texttt{conv}}
\newcommand{\iset}[2]{\mathcal{#1}_{#2}}
\newtheorem{assumption}{Assumption}

\newcommand{\fullversion}[2]{#2}

\begin{document}

\newcommand\relatedversion{}

\makeatletter
\def\thanks#1{\protected@xdef\@thanks{\@thanks
        \protect\footnotetext{#1}}}
\makeatother

\title{Stochastic Submodular Maximization \\via Polynomial Estimators}
%
%
\author{G\"{o}zde \"{O}zcan\inst{}\orcidID{0000-0002-2957-6893}
\and Stratis Ioannidis\inst{}\orcidID{0000-0001-8355-4751}}

\authorrunning{\replaced{G. \"{O}zcan and S. Ioannidis}{Anonymous Authors}}

%
\institute{Electrical and Computer Engineering Department, Northeastern University, Boston MA 02115, USA\\
\email{\{gozcan, ioannidis\}@ece.neu.edu}}

\maketitle
%
    
\begin{abstract}
In this paper, we study stochastic submodular maximization problems with general matroid constraints, that naturally arise in online learning, team formation, facility location, influence maximization, active learning and sensing objective functions. In other words, we focus on maximizing submodular functions that are defined as expectations over a class of submodular functions with an unknown distribution. We show that for monotone functions of this form, the stochastic continuous greedy algorithm \cite{mokhtari2020stochastic} attains an approximation ratio (in expectation) arbitrarily close to $(1-1/e) \approx 63\%$ using a polynomial estimation of the gradient. We argue that using this polynomial estimator instead of the prior art that uses sampling eliminates a source of randomness and experimentally reduces execution time.



\end{abstract}

\section{Introduction}
Submodular maximization is a true workhorse of data mining, arising in settings as diverse as  hyper-parameter optimization \cite{susubmodular}, feature compression \cite{bateni2019categorical}, text classification \cite{lei2019discrete}, and influence maximization \cite{goyal2011data, kempe2003maximizing}. Many of these interesting problems as well as variants can be cast as maximizing a submodular set function $f(S)$, defined over sets $S\subseteq V$ for some ground set $V$,  subject to a matroid constraint. Despite the NP-hardness of these problems, the so-called \emph{continuous-greedy} (CG) algorithm~\cite{calinescu2011maximizing}, can be used to construct a $1-1/e$-approximate solution in polynomial time. Interestingly, the solution is generated by first transferring the problem to the continuous domain, and solving a continuous optimization problem via gradient techniques. The  solution to this continuous optimization problem is subsequently rounded (via techniques such as pipage rounding \cite{ageev2004pipage} and swap rounding \cite{calinescu2011maximizing}), to produce an integral solution within a $1-1/e$ factor from the optimal.
 The continuous optimization problem solved by the CG  algorithm amounts to maximizing so-called  \emph{multilinear relaxation} of the original, combinatorial submodular objective.
In short, the multilinear relaxation  of a submodular function $f(S)$ is its expectation assuming its input $S$ is generated via independent Bernoulli trials, and is typically computed via sampling \cite{calinescu2011maximizing, vondrak2008optimal}.

Recently, a series of papers have studied an interesting variant called the \emph{stochastic submodular optimization} setting~\cite{asadpour2008stochastic, zhang2022stochastic, chen2018online,karimi2017stochastic,hassani2017gradient,mokhtari2020stochastic}. In this setting, the submodular objective function to be optimized is assumed to be of the form of an expectation, i.e., $f(S)=\mathbb{E}_{z\sim P}[f_z(S)]$, where $z$ is a random variable. Moreover, the optimization algorithm does not have access to the a function oracle (i.e., cannot compute the function itself). Instead it can only sample a random instantiation of $f_z(\cdot)$, different each time. This setting is of course of interest when the system or process that $f$ models is inherently stochastic (e.g., involves a system dependent on, e.g., user behavior or random arrivals) and the distribution governing this distribution is not a priori known. It is also of interest when the support of distribution $P$ is very large, so that the expectation cannot be completed efficiently. A classic example of the latter case is influence maximization (c.f.~Sec.~\ref{sec:probdef}), where the expectation  $f(S)$ cannot be computed efficiently or even in a closed form, even though samples $z\sim P$ can be drawn.

Interestingly, the fact that the classic continuous greedy algorithm operates in the continuous domain gives rise to a \emph{stochastic continuous greedy} (SCG) method for tackling the stochastic optimization problem~\cite{mokhtari2020stochastic}.  In a manner very similar to stochastic gradient descent, the continuous greedy algorithm can be modified to use \emph{stochastic gradients}, i.e., random variables whose expectations equal the gradient of the multilinear relaxation. In practice, these are computed by sampling \emph{two random variables in tandem}: $z\sim P$, which is needed to generate a random instance $f_z$, and $S$, the random input needed to compute the multilinear relaxation. As a result, the complexity of the SCG algorithm depends on the variance due to \emph{both} of these two variables.

We make the following contributions:
\begin{itemize}
    \item We use polynomial approximators, originally proposed by \"Ozcan et al.~\cite{ozcan2021submodular}, to reduce the variance of the stochastic continuous greedy algorithm. In particular, we eliminate one of the two sources of randomness of SCG, namely, sampling $S$. We do this by replacing the sampling estimator by a deterministic estimator constructed by approximating each $f_z(\cdot)$ with a polynomial function.
    \item We show that doing so \emph{reduces the variance} of the gradient estimation procedure used by SCG, but introduces a \emph{bias}. We then characterize the performance of SCG in terms of both the (reduced) variance and new bias term.
    \item We show that for several interesting stochastic submodular maximization problems, including influence maximization, the bias can be well-controlled, decaying exponentially with the degree of our polynomial approximators.
    \item Finally, we illustrate the advantage of our approach experimentally, over both synthetic and real-life datasets.
\end{itemize}

\section{Related Work}
  While  submodular optimization problems are generally NP-hard, the celebrated greedy algorithm \cite{nemhauser1978analysis} attains a $(1-1/e)$ approximation ratio for  submodular maximization subject to uniform matroids and a $1/2$ approximation ratio for general matroid constraints. As discussed in the introduction, the  continuous greedy algorithm \cite{calinescu2011maximizing} restores the $(1-1/e)$ approximation ratio by lifting the discrete problem to the continuous domain via the multilinear relaxation. 

Stochastic submodular maximization, in which the objective is expressed as an expectation, has gained a lot of interest in the recent years \cite{asadpour2008stochastic, zhang2022stochastic, chen2018online}. Karimi et al. \cite{karimi2017stochastic} use a concave relaxation method that achieves the $(1-1/e)$ approximation guarantee, but only  for the class of submodular coverage functions. Hassani et al.~\cite{hassani2017gradient} provide projected gradients methods for the general case of stochastic submodular problems that achieve $1/2$ approximation guarantee.  Mokhtari et al. \cite{mokhtari2020stochastic} propose stochastic  conditional gradient methods for solving both minimization and maximization  stochastic submodular optimization problems. Their method for maximization, Stochastic Continous Greedy (SCG) can be interpreted as a stochastic variant of the continuous greedy algorithm \cite{vondrak2008optimal, calinescu2011maximizing} and achieves a tight $(1-1/e)$ approximation guarantee for monotone and submodular functions. 

Our work builds upon and relies on the approach by  \"{O}zcan et al.~\cite{ozcan2021submodular}, who studied ways of accelerating the computation of gradients via a polynomial estimator. Extending on the work of Mahdian et al.~\cite{mahdian2020kelly},  \"{O}zcan et al. show that submodular functions that can be written as compositions of (a) an analytic function and (b) a multilinear function can be arbitrarily well approximated via Taylor polynomials; in turn, this gives rise to a method for approximating their multilinear relaxation in a closed form, without sampling. We leverage this method in the context of stochastic submodular optimization, showing that it can also be applied in combination with SCG of Mokhtari et al.~\cite{mokhtari2020stochastic}: this eliminates one of the two sources of randomness, thereby reducing variance at the expense of added bias. From a technical standpoint, this requires controlling the error introduced by the bias of the polynomial estimator, while simultaneously accounting for the variance inherent in SCG, due to sampling instances.   

\section{Technical Preliminary}

\subsubsection{Submodularity and Matroids.}\label{sec:submat}
Given a ground set $V = \{1, \ldots, n\}$ of $n$ elements, a set function $f:2^V\rightarrow\reals_+$ is submodular if and only if $f(B \cup \{e\}) - f(B) \leq f(A\cup \{e\}) - f(A)$, for all $A\subseteq B\subseteq V$ and $e\in V$. Function $f$ is \emph{monotone} if $f(A)\leq f(B)$, for every $A\subseteq B$.

\noindent \textbf{Matroids.} Given a ground set $V$, a matroid is a pair $\mathcal{M}=(V, \mathcal{I})$, where $\mathcal{I}\subseteq 2^V$ is a collection of \emph{independent sets}, for which the following hold: 
(a) if $B\in \mathcal{I}$ and $A \subset B$, then $A \in \mathcal{I}$, and (b)
  if $A, B\in \mathcal{I}$ and $|A|< |B|,$ there exists $x \in B\setminus A$ s.t. $A\cup\{x\}\in \mathcal{I}$.
 The \emph{rank} of a matroid $r_{\mathcal{M}}(V)$ is the largest cardinality of its elements, i.e.:
  $  r_{\mathcal{M}}(V) = \max\{|A|: {A}\in\mathcal{I}\}.$
We introduce two examples of matroids:
\begin{enumerate}
    \item \textbf{Uniform Matroids.} The uniform matroid with cardinality $k$ is $\mathcal{I}=\{S\subseteq V, \, |S|\leq k\}$.
    \item \textbf{Partition Matroids.} Let $\mathcal{B}_1,\ldots, \mathcal{B}_m\subseteq V$ be a partitioning of $V$, i.e., $ \bigcap_{\ell=1}^m\mathcal{B}_\ell =\emptyset$ and $\bigcup_{\ell=1}^m\mathcal{B}_\ell = V$. Let also $k_\ell\in \mathbb{N}, \ell=1,\ldots,m$, be a set of cardinalities.   A partition matroid is defined as $\mathcal{I}=\{S\subseteq 2^V \, \mid  \, |S\cap \mathcal{B}_\ell|\leq k_{\ell}, \text{ for all } \ell=1,\ldots, m\}.$  
\end{enumerate}

\subsection{Problem Definition}\label{sec:probdef}
In this work, we focus on \textit{discrete  stochastic submodular maximization} problems. More specifically, we consider set function $f: 2^V \rightarrow \mathbb{R}_+$ of the form:
 $   f(S) = \mathbb{E}_{z \sim P}[f_z(S)],$ $S \subseteq V,$
where  $z$ is the realization of the random variable $Z$ drawn from a distribution $P$ over a probability space $(V_z,P)$. For each realization of $z \sim P$, the set function $f_z: 2^V \rightarrow \mathbb{R}_+$ is monotone and submodular. Hence, $f$ itself is monotone and submodular. The objective is to maximize $f$ subject to some constraints (e.g., cardinality or matroid constraints) by only accessing to i.i.d. samples of $f_{z \sim P}$. In other words, we wish to solve:
\begin{equation} \label{prob:stochsubmax}
    \max_{S \in \mathcal{I}} f(S) = \max_{S \in \mathcal{I}} \mathbb{E}_{z \sim P}[f_z(S)],
\end{equation}
where  $\mathcal{I}$ is a general matroid constraint.

Stochastic submodular maximization problems are of interest in the absence of the oracle that provides the exact value of $f(S)$: one can only access $f_z(S)$, for random instantiations $z\sim P$.  A well-known motivational example is contagion propagation in a network (a.k.a., the influence maximization problem \cite{kempe2003maximizing}). Given a graph with node set $V$, reachability of nodes from seeds are determined by sampling sub-graph $G=(V, E)$, via, e.g., the Independent Cascade or the Linear Threshold model~\cite{kempe2003maximizing}. The random edge set, in this case, plays the role of $z$, and the distribution over graphs the role of $P$.  The function $f_z(S)$ represents the ratio of nodes reachable from the seeds $S$ under the connectivity induced by edges $E$ in this particular realization of $z$. 
The goal is to select seeds $S$ that maximize $f(S) = \mathbb{E}_{z \sim P}[f_z(S)]$; both $f$ and $f_z$ are monotone submodular functions; however computing $f$ in a closed form is hard, and $f(\cdot)$ can only be accessed through random instantiations of $f_z(\cdot)$.


\subsection{Change of Variables and Multiliear Relaxation}  
There is a 1-to-1 correspondence between a binary vector $\vc{x}\in \{0,1\}^{n}$ and its support $S=\texttt{supp}(\vc{x})$. Hence, a set function $f: 2^V \rightarrow \reals_+$ can be interpreted as $f: \{0,1\}^n \rightarrow \reals_+$ via: 
$f(\vc{x}) \triangleq f(\texttt{supp}(\vc{x}))$ for $\vc{x} \in \{0,1\}^n$. We adopt this convention for the remainder of the paper. 
We also treat matroids as subsets of $\{0,1\}^n$, defined consistently with this change of variables via $\mathcal{M}=\{\vc{x}\in\{0,1\}^n: \supp(\vc{x})\in \mathcal{I}\}.$ For example, a partition matroid is: 
$\mathcal{M} = \textstyle\left\{\vc{x} \in \{0,1\}^n\,\mid \bigcap_{\ell=1}^m  \left(\sum_{i\in B_\ell} x_i\leq k_\ell\right)\right\}.$
The \emph{matroid polytope} $\mathcal{C}\subseteq [0,1]^{n}$ is the convex hull of matroid $\mathcal{M}$, i.e., $\mathcal{C} = \cvx(\mathcal{M}).$

We define the \emph{multilinear relaxation} of $f$ as:
\begin{align}\label{eq: multilinear}
\begin{split}
    G(\mathbf{y}) & = \mathbb{E}_{S \sim \mathbf{y}}[f(S)] = \sum_{S \subseteq V} f(S) \prod_{i \in S} y_i \prod_{j \notin S} (1 - y_j)\\&=\mathbb{E}_{\mathbf{x} \sim \mathbf{y}}[f(\mathbf{x})]=\sum_{\mathbf{x}\in \{0,1\}^n} f(\mathbf{x}) \prod_{i\in V}y_i^{x_i}(1-y_i)^{(1-x_i)},  \quad \text{for}~\mathbf{y} \in [0, 1]^n.
    \end{split}
\end{align}
In other words, $G:[0,1]^n\to\reals_+$ is the expectation of $f$, assuming that $S$ is random and generated from independent Bernoulli trials: for every $i\in V$, $P(i\in S)=y_i$. 
The multilinear relaxation of $f$ satisfies several properties. First, it is indeed a relaxation/extension of $f$ over the (larger) domain $[0,1]^n$: for $\mathbf{x}\in \{0,1\}^n$, $G(\mathbf{x})=f(\mathbf{x})$, i.e., $G$ agrees with $f$ on integral inputs. Second, it is \emph{multilinear} (c.f.~Sec.~\ref{sec:multilinear}), i.e., affine w.r.t.~any single coordinate $y_i$, $i\in V$, when keeping all other coordinates $\mathbf{y}_{-i}=[y_j]_{j\neq i}$ fixed. 
Finally, in the context of stochastic submodular optimization,  it is an expectation that involves \emph{two sources of randomness}: (a) $z\sim P$, i.e., the random instantiation of the objective, \emph{as well as } (b) $\mathbf{x}\sim \mathbf{y}$, i.e., the independent sampling of the Bernoulli variables (i.e., the set $S$). In particular,  we can write: 
\begin{align}
    G(\mathbf{y}) = \mathbb{E}_{z\sim P}[G_z(\mathbf{y})],~\text{where}~G_z(\mathbf{y}) = \mathbb{E}_{\mathbf{x}\sim \mathbf{y}}[f_z(x)]~\text{is the multilinear relaxation of}~ f_z(\cdot).  
\end{align}



\subsection{Stochastic Continuous Greedy Algorithm}
The stochastic nature of the set function $f(S)$ requires the use the \emph{Stochastic Continuous Greedy (SCG)} algorithm \cite{mokhtari2020stochastic}. This is a stochastic variant of the  continuous greedy algorithm (method) \cite{vondrak2008optimal}, to solve (\ref{prob:stochsubmax}). The SCG algorithm uses a common averaging technique in stochastic optimization and computes the estimated gradient $\mathbf{d}_t$ by the recursion
\begin{equation}\label{eq:avgGrad}
    \mathbf{d}_t = (1 - \rho_t)\mathbf{d}_{t-1} + \rho_t \nabla G_{z_t} (\mathbf{y}_t),
\end{equation}
where $\rho_t$ is a positive step size and the algorithm initially starts with $\mathbf{d}_0 = \mathbf{y}_0 = \mathbf{0}$. Then, it proceeds in iterations, where in the $t$-th iteration it finds a feasible solution as follows
\begin{equation}
    \mathbf{v}_t \in \arg\max_{\mathbf{v} \in \mathcal{C}} \{\mathbf{d}_t^T\mathbf{v}\},
\end{equation}
where $\mathcal{C}$ is the matroid polytope (i.e., convex hull) of matroid $\mathcal{M}$.
After finding the ascent direction $\mathbf{v}_t$, the current solution $\mathbf{y}_t$ is updated as
\begin{equation}
    \mathbf{y}_{t+1} = \mathbf{y}_t + \frac{1}{T}\mathbf{\replaced{v}{y}}_t, 
\end{equation}
where $1/T$ is the step size. The steps of the stochastic continuous greedy algorithm are outlined in Algorithm~\ref{alg: SCG}.
The (fractional) output of Algorithm~\ref{alg: SCG} is within a $1-1/e$ factor from the optimal solution to Problem~(\ref{prob:stochsubmax}) (see Theorem~\ref{thm:main} below). This fractional solution can subsequently be rounded in polynomial time to produce a solution  with the same approximation guarantee w.r.t.~to Problem~(\ref{prob:stochsubmax}) using, e.g., either the pipage rounding \cite{ageev2004pipage} or the swap rounding \cite{chekuri2010dependent} methods.
\begin{algorithm}[!t]
    \caption{Stochastic Continuous Greedy (SCG)}\label{alg: SCG}
    \textbf{Require:} Step sizes $\rho_t > 0$. Initialize $\mathbf{d}_0 = \mathbf{y}_0 = \mathbf{0}.$
    \begin{algorithmic}[1] 
            \For{$t = 1, 2, \ldots, T$}
                \State Compute $\mathbf{d}_t = (1 - \rho_t) \mathbf{d}_{t-1} + \rho_t \nabla G_{z_t}(\mathbf{y}_t)$;
                \State Compute $\mathbf{v}_t \in \arg\max_{\mathbf{v} \in \mathcal{C}} \{\mathbf{d}_t^T \mathbf{v}\}$;
                \State Update the variable $\mathbf{y}_{t+1} = \mathbf{y}_t + \frac{1}{T} \mathbf{v}_t$;
            \EndFor
    \end{algorithmic}
\end{algorithm}
\subsubsection{Sample Estimator.}
The gradient $\nabla G_{z_t}$ is needed to perform step \eqref{eq:avgGrad}; computing it directly via Eq.~\eqref{eq: multilinear}. requires exponentially many calculations. Instead, both Calinescu et al. \cite{calinescu2011maximizing} and Mokhtari et al. \cite{mokhtari2020stochastic} estimate it via \emph{sampling}. 
In particular, due to multilinearity (i.e., the fact that $G_z$ is affine w.r.t. a coordinate $x_i$, we have: 
\begin{equation}\label{eq:partialGrad}
    \frac{\partial G_z(\mathbf{y})}{\partial x_i} = G_z ([\mathbf{y}]_{+i}) - G_z ([\mathbf{y}]_{-i}),\quad\text{for all}~i\in V,
\end{equation}
where $[\mathbf{y}]_{+i}$ and $[\mathbf{y}]_{-i}$ are equal to the vector $\mathbf{y}$ with the $i$-th coordinate set to $1$ and $0$, respectively. The gradient of $G$ can thus be estimated by (a) producing $N$ random samples $\mathbf{x}^{(l)}$, for $l \in \{1, \ldots, N\}$ of the random vector $\mathbf{x}$, and (b) computing the empirical mean of the r.h.s. of (\ref{eq:partialGrad}), yielding
\begin{equation}\label{eq:sampleEst}
    \frac{\partial \widehat{G_{z}(\mathbf{y})}}{\partial x_i} = \frac{1}{N} \sum_{l=1}^N \left(f_z ([\mathbf{x}^{(l)}]_{+i}) - f_z ([\mathbf{x}^{(l)}]_{-i})\right), \quad\text{for all}~i\in V.
\end{equation}

Mokhtari et al. \cite{mokhtari2020stochastic} make the following assumptions:
\begin{assumption} \label{asm:monSub}
    Function $f: \{0, 1\}^n \rightarrow \mathbb{R}_+$ is monotone and submodular.
\end{assumption}

\begin{assumption} \label{asm:boundedNorm}
    The Euclidean norm of the elements in the constraint set $\mathcal{C}$ are uniformly bounded, i.e., for all $\mathbf{y} \in \mathcal{C}$, there exists a $D$ s.t. $\|\mathbf{y}\| \leq D.$
\end{assumption}

Under these assumptions, SCG combined with the sampling estimator in Eq.~\eqref{eq:partialGrad}, yields the following guarantee:
\begin{theorem} \label{thm:theirs}
    [Mokhtari et al. \cite{mokhtari2020stochastic}] Consider Stochastic Continuous Greedy (SCG) outlined in Algorithm~\ref{alg: SCG}, with $\nabla G_{z_t}(\mathbf{y}_t)$ replaced by $\nabla \widehat{G_{z_t}(\mathbf{y}_t)}$ given by (\ref{eq:sampleEst}). Recall the definition of the multilinear extension function $G$ in \eqref{eq: multilinear} and set the averaging parameter as $\rho_t = 4/(t+8)^{2/3}$. If Assumptions~\ref{asm:monSub}~\&~\ref{asm:boundedNorm} are satisfied, then the iterate $\mathbf{y}_T$ generated by SCG satisfies the inequality
    \begin{equation}
        \mathbb{E}\left[G(\mathbf{y}_T)\right] \geq (1-1/e)OPT - \frac{15DK}{T^{1/3}} - \frac{f_{\max}rD^2}{2T},
    \end{equation}
    where $OPT = \max_{\mathbf{y} \in \mathcal{C}} G(\mathbf{y})$ and $K = \max\{3\|\nabla G(\mathbf{y}_0) - \mathbf{d}_0\|, 4 \sigma + \sqrt{3r} f_{\max}D\},$ where $D$ is the diameter of the convex hull $\mathcal{C}$, $f_{\max}$ is the maximum marginal value of the function $f$, i.e., $f_{\max} = \max_{i \in \{1, \ldots, n\}} f(\{i\})$,  $r$ is the rank of the matroid $\mathcal{I}$, and
    $\sigma^2 =\sup_{\mathbf{y}\in \mathcal{C}} \mathbb{E}\left[\|\widehat{\nabla G_z(\mathbf{y}) } - G(\mathbf{y})  \|\right],$
    where $\widehat{\nabla G_z}$ is the sample estimator given by Eq.~\eqref{eq:sampleEst}.
\end{theorem}
Thus, by appropriately setting the number of iterations $T$, we can produce a solution that is arbitrarily close to $1-1/e$ from the optimal (fractional) solution. Again, this can be subsequently rounded (see, e.g., \cite{ageev2004pipage,calinescu2011maximizing}) to produce an integer solution with the same approximation guarantee.
It is important to note that the number of steps required depends on $\sigma^2$, which is a (uniform over $\mathcal{C}$) bound on the variance of the estimator given by Eq.~\eqref{eq:sampleEst}. This variance contains \emph{two sources of randomness}, namely $z\sim P$, the random instantiation, and $\mathbf{x}\sim\mathbf{y}$, as multiple such integer vectors/sets are sampled in Eq~\eqref{eq:sampleEst}. In general, the variance will depend on the number of samples $N$ in the estimator, and will be bounded (as $G$ is bounded).\footnote{For example, even for $N=1$, the submodularity of $f_z$ and Eq.~\eqref{eq:partialGrad} imply that $\sigma^2\leq 2 {n}{\max_{j \in [n]} \mathbb{E}[f_z(\{j\})^2]}$ \cite{mokhtari2020stochastic}, though this bound is loose/a worst-case bound.} 
\subsection{Multilinear Functions and the Multilinear Relaxation of a Polynomial}\label{sec:multilinear}
Recall that a \emph{polynomial} function $p: \reals^n \rightarrow \mathbb{R}$ can be written as a linear combination of several monomials, i.e.,
\begin{equation} \label{eq:polyFormat}
     p(\mathbf{y}) = c_0+\sum_{\ell \in \mathcal{I}} c_{\ell} \prod_{i \in \iset{J}{\ell}} y_i^{k_i^{\ell}},
\end{equation}
where $c_{\ell}~\in~\reals$ for $\ell$ in some index set $\mathcal{I}$, subsets $\iset{J}{\ell}~\subseteq~V$ determine the terms of each monomial, and 
, and $\{k_i^{\ell}\}_{i \in \iset{J}{\ell}} \subset \mathbb{N}$ are natural exponents. W.l.o.g. we assume that $k_i^{\ell} \geq 1$ (as variables with zero exponents can be ommited). The degree of the monomial indexed by $\ell\in\iset{I}{}$ is $k^\ell=\sum_{i\in \iset{J}{\ell}} {k_i^\ell}$, and the degree of polynomial $p$ is $\max_{\ell\in \iset{I}{}} k^\ell$, i.e., the largest degree across monomials.

A function $f: \reals^N \rightarrow \reals$ is \emph{multilinear} if it is affine w.r.t.~each of its coordinates \cite{broida1989comprehensive}. 
Alternatively, multilinear functions are polynomial functions in which the degree of each variable in a monomial is at most $1$; that is, multilinear functions can be written as:
\begin{equation} \label{eq:multi}
    f(\mathbf{y}) = c_0'+\sum_{\ell \in \mathcal{I}} c_{\ell}' \prod_{i \in \iset{J}{\ell}} y_i,
\end{equation}
where $c_{\ell}~\in~\reals$ for $\ell$ in some index set $\mathcal{I}$, and subsets $\iset{J}{\ell}~\subseteq~V$, again determining monomials of degree \emph{exactly equal to} ${|\iset{J}{\ell}|}.$ 
Given a polynomial $p$ defined by the parameters in Eq.~(\ref{eq:polyFormat}), let  
\begin{equation} \label{eq:multiFormat}
     \dot{p}(\mathbf{y}) = c_0+\sum_{\ell \in \mathcal{I}} c_{\ell} \prod_{i \in \iset{J}{\ell}} y_i,
\end{equation}
be the multilinear function resulting from  $p$, \emph{by replacing all its exponents $k_i^{\ell} \geq 1$ with $1$}. We call this function the \emph{multilinearization} of $p$. The multilinearization of $p$ is inherently linked to its multilinear relaxation:
\begin{lemma}[\"Ozcan et al.~\cite{ozcan2021submodular}] \label{lem:relaxation_of_multi}
Let $p: [0, 1]^n \rightarrow \mathbb{R}$ be an arbitrary polynomial and let $\dot{p}:\reals^n \rightarrow \reals_+$ be  its multilinearization, given by Eq.~\eqref{eq:multiFormat}. Let $\vc{x} \in \{0, 1\}^n$ be a random vector of independent Bernoulli coordinates parameterized by $\vc{y}\in~[0, 1]^n$. Then, $\mathbb{E}_{\vc{x}\sim\vc{y}}[p(\vc{x})] = \mathbb{E}_{\vc{x}\sim\vc{y}}[\dot{p}(\vc{x})] =\dot{p}(\vc{y}).$
\end{lemma}
\begin{proof}
Observe that 
 $   \dot{p}(\mathbf{x}) = p(\mathbf{x})$, for all $\mathbf{x} \in \{0, 1\}^n.$ 
This is precisely because $x^{k} = x$ for  $x \in \{0, 1\}$ and all $k \geq 1$.  The first equality therefore follows. On the other hand, $\dot{p}(\mathbf{x})$ is the multilinear function given by Eq.~\eqref{eq:multiFormat}. Hence
 $   \mathbb{E}_{\vc{x}\sim\vc{y}}[\dot{p}(\vc{x})] = \mathbb{E}_{\vc{x}\sim\vc{y}}\left[ c_0+\sum_{\ell \in \mathcal{I}} c_{\ell} \prod_{i \in \iset{J}{\ell}} x_i\right] = c_0+\sum_{\ell \in \mathcal{I}} \mathbb{E}_{\vc{x}\sim\vc{y}}\left[ \prod_{i \in \iset{J}{\ell}} x_i\right]=c_0+\sum_{\ell \in \mathcal{I}}  \prod_{i \in \iset{J}{\ell}}\mathbb{E}_{\vc{x}\sim\vc{y}}\left[ x_i\right]=\dot{p}(\mathbf{y}),$
where the second to last equality holds by the independence across $x_i$, $i\in V$.
\end{proof}
An immediate consequence of this lemma is that the multilinear relaxation of any polynomial function can be computed \emph{without sampling}, by simply computing its multilinearization. This is of particular interest of course for submodular functions that are themselves polynomials (e.g., coverage functions~\cite{karimi2017stochastic}). \"Ozcan et al.~extend this to submodular functions that can be written as compositions of a scalar and a polynomial function, by approximating the former via its Taylor expansion. We extend  and generalize this to the case of stochastic submodular functions, so long as the latter can be approximated arbitrarily well by polynomials.  

\section{Main Results}

\subsection{Polynomial Estimator}
To leverage Lem.~\ref{lem:relaxation_of_multi} to the case of stochastic submodular functions, we make the following  assumption:
\begin{assumption} \label{asmp:boundedPoly}
    For all $z 
    \in V_z$, there exists a sequence of polynomials $\{\hat{f}_z^L\}_{L=1}^{\infty}$, $\hat{f}_z^L: \mathbb{R}^n \rightarrow \mathbb{R}$ such that  
    $\lim_{L \rightarrow \infty} |f_z(\mathbf{x}) - \hat{f}_z^L(\mathbf{x})| = 0,$ uniformly over $\mathbf{x} \in \{0, 1\}^n,$
    i.e. there exists $\varepsilon_z (L) \geq 0$ such that $\lim_{L \rightarrow \infty} \varepsilon_z (L) = 0$ and $|f_z(\mathbf{x}) - \hat{f}_z^L(\mathbf{x})| \leq \varepsilon_z(L)$,  for all $\mathbf{x} \in \{0, 1\}^n$.
\end{assumption}


 In other words, we assume that we can asymptotically approximate every function $f_z$ with a polynomial arbitrarily well. Note that there already exists a polynomial function that approximates each $f_z$ \emph{perfectly} (i.e., $\epsilon_z=0$), namely, its multilinear relaxation $G_z$. However, the number of terms in this polynomial is exponential in $n$. In contrast, Asm.~\ref{asmp:boundedPoly} requires exact recovery only asymptotically. In many cases, this allows us to construct polynomials with only a handful (i.e., polynomial in $n$) terms, that can approximate $f_z$. We will indeed present such polynomials for several applications of interest in Section~\ref{sec: examples}.
 Armed with this assumption, we define an estimator $\widehat{\nabla G_z^L}$ of the gradient  of the multilinear relaxation $G$ as follows:
\begin{equation}
    \begin{split}
        \frac{\widehat{{\partial G_z^L}}}{\partial y_i}\big|_{\vc{y}} &\equiv \mathbb{E}_{\vc{x}\sim\vc{y}}[\hat{f}_z^{L}([\vc{x}]_{+i})] - \mathbb{E}_{\vc{x}\sim\vc{y}}[\hat{f}_z^{L}([\vc{x}]_{-i})]  
         \stackrel{\text{Lem.}~\ref{lem:relaxation_of_multi}}{=} \dot{\hat{f}}_z^{L}([\vc{y}]_{+i}) - \dot{\hat{f}}_z^{L}([\vc{y}]_{-i}), \text{ for all $i \in V$}.\label{eq: poly_estimator}
    \end{split}
\end{equation}
In other words, our estimator is constructed by replacing the multilinear relaxation $G_z$ in Eq.~\eqref{eq:partialGrad} with the multilinear relaxation of the approximating polynomial $\hat{f}_z$. In turn,  by Lem.~\ref{lem:relaxation_of_multi}, \emph{the latter can be computed deterministically  (without any sampling of the Bernoulli variables $\mathbf{x}\sim \mathbf{y}$)}, in closed form: the latter is given by the multilinearization  $\dot{\hat{f}}_z^L$ of polynomial ${\hat{f}}_z^L$.

Nevertheless, our deterministic estimator given by Eq.~\eqref{eq: poly_estimator} has a \emph{bias}, precisely because of our approximation of $f_z$ via the polynomial $\hat{f}_z^L$. We characterize this bias via the following lemma:
\begin{lemma} \label{lem:gradientBias}
Assume that function $f_z$ satisfies Asm.~\ref{asmp:boundedPoly}. Let $\nabla G_z$ be the unbiased stochastic gradient for a given $f_z$ and let $\widehat{\nabla G_z^L}$ be the estimator of the multilinear relaxation given by \eqref{eq: poly_estimator}. Then, 
 $   \big\|\nabla G_z(\vc{y}) - \widehat{\nabla G_z^L}(\vc{y})\big\|_2 \leq 2\sqrt{n}\varepsilon_z (L),$ for all $\mathbf{y} \in \mathcal{C}$.
\end{lemma}

The proof can be found in App.~\ref{app:proof_gradBiasLemma}\deleted{of the supplement}. Hence, we can approximate $\nabla G$ arbitrarily well, uniformly over all $\vc{x}\in[0,1]^n$.
We can thus use our estimator in the SCG algorithm instead of of the sample estimator of the gradient (Eq.~\eqref{eq:sampleEst}). 
We prove that this yields the following guarantee:
\begin{theorem}\label{thm:main}
    Consider Stochastic Continuous Greedy (SCG) outlined in Algorithm~\ref{alg: SCG}. Recall the definition of the multilinear extension function $G$ in \eqref{eq: multilinear}
    . If Asm.~\ref{asm:monSub}  
    is satisfied and $\rho_t = 4/(t+8)^{2/3}$
    , then the objective function value for the iterates generated by SCG satisfies the inequality
        \begin{align*}
            \mathbb{E}[G(\mathbf{y}_T)] \geq (1-1/e) OPT - \frac{15 D K}{T^{1/3}} - \frac{f_{\max} r D^2}{2T},
        \end{align*}
    where $K = \max\{3\|\nabla G(\mathbf{y}_0 - \mathbf{d}_0)\|^2, \sqrt{16\sigma_0^2 + 224\sqrt{n}\varepsilon (L)} + 2\sqrt{r}f_{\max}D\},$ $OPT = \max_{\mathbf{y} \in \mathcal{C}} G(\mathbf{y})$, $r$ is the rank of the matroid $\mathcal{I}$, $\varepsilon(L)=\mathbb{E}_{z\sim P}[\varepsilon_z(L)]$,  $f_{\max}$ is the maximum marginal value of the function $f$, i.e., $f_{\max} = \max_{i \in \{1, \ldots, n\}} f(\{i\})$, and 
    $\sigma_0^2 = \sup_{\mathbf{y}\in \mathcal{C}} \mathbb{E}_{z\sim P}\left[\left\|\nabla G(\mathbf{y}) - \nabla G_{z}(\mathbf{y})\right\|^2 \right].$
    \end{theorem}
The proof of the theorem can be found
 \fullversion{in App.~\ref{app: proofMainThm} \deleted{of the supplementary material.}}{in App.~\ref{app: proofMainThm} \deleted{of the supplementary material.}} Our proof follows the main steps of \cite{mokhtari2020stochastic}
 , using however the bias guarantee from Lem.~\ref{lem:gradientBias}; to do so, we need to deal with the fact that our estimator is not unbiased, but also that stochasticity is still present (as variables $z$ are still sampled randomly). This is also reflected in our bound, that contais both a bias term (via $\varepsilon(L)$) and a variance term (via $\sigma_0$).
 
 Comparing our guarantee to Thm.~\ref{thm:theirs}, we observe two main differences. On one hand, we have replaced the uniform bound of the variance $\sigma^2$ with the smaller quantity $\sigma^2_0$: the latter is quantifying the gradient variance w.r.t.~$z$, and is thus smaller than $\sigma$, that depends on the variance of  \emph{both} $z$ \emph{and} $\mathbf{x}\sim \mathbf{y}$. Crucially, $\sigma^2_0$ is an ``inherent'' variance, \emph{independent of the gradient estimation process}: it is the variance due to the  randomness $z$, which is inherent in how we access our  stochastic submodular objective and thus cannot be avoided.  On the other hand, this variance reduction comes at the expense of introducing a bias term. This, however, can be suppressed via Asm.~\ref{asmp:boundedPoly}; as we discuss in the next section, for several problems of interest, this can be made arbitrarily small using only a polynomial number of terms in $\hat{f}^L_z$.

\section{Problem Examples}\label{sec: examples}
In this section, we list several problems that can be tackled through our approach, also summarized in Tab.~\ref{table: problems}; these are similar to the problems considered by \"Ozcan et al.~\cite{ozcan2021submodular}, but cast into the stochastic submodular optimization setting. All problems correspond to trivially bounded variances $\sigma_0^2$ (again, because functions $f_z$ are bounded); we thus focus on determining their bias $\epsilon(L)$. For space reasons, we report Cache Networks (CN) in Table~\ref{table: problems}, but provide details for it in the \deleted{supplement }\deleted{(}App.~\ref{app:customBias}\deleted{)}.

\begin{table*}[ht] 
\caption{Summary of problems satisfying Asm.~\ref{asm:monSub}\&~\ref{asmp:boundedPoly}.}
\centering \label{table: problems}
\resizebox{\textwidth}{!}{
    \begin{tabular}{ |c|c|c|c|c|c| } 
     \cline{2-6}
     \multicolumn{1}{c}{} 
     & \multicolumn{1}{|c|}{\thead{Input}} 
     & \multicolumn{1}{|c|}{\thead{$g_z: \{0, 1\}^{|V|} \rightarrow [0, 1]$\\
                                   $\vc{x} \rightarrow g_z(\vc{x})$}} 
     & \multicolumn{1}{|c|}{\thead{$f_z:\{0, 1\}^{|V|} \rightarrow \reals_+$\\
                                   $\vc{x} \rightarrow f_z(\vc{x})$}} 
     & \multicolumn{1}{|c|}{\thead{$\hat{f}_z^L:\{0, 1\}^{|V|} \rightarrow \reals_+$\\
                                   $\vc{x} \rightarrow \hat{f}_z^L(\vc{x})$}}
     & \multicolumn{1}{|c|}{\thead{Bias\\
                                   $\varepsilon(L)$}}\\
     \hline
     \thead{SM} 
     & \makecell{Weighted bipartite graph \\
                $G = (V \cup P)$ weights $\vc{r}_z\in\reals_+^{n}$, \\
                and $\sum_{i=1}^n r_{i, z}=1$
                } 
     & $\sum_{i \in V\cap P_j} r_{i, z}x_i$ 
     & \makecell{$\sum_{j=1}^J h\left(g_z(\mathbf{x})\right)$,\\ where\\
     $h(s) = \log(1+s)$} 
     & \makecell{$\hat{h}^L(g_z(\mathbf{x}))$, \\
     where $\hat{h}^L$ is 
     Eq.~\eqref{eq: taylor_f_iL}}
     & $\frac{1}{(L+1) 2^{L+1}}$ \\
     \hline
     \thead{IM} 
     & \makecell{Instances $G = (V, E)$\\
                 of a directed graph, \\
                 partitions $P_{v}^z \subset V$
                 } 
     & $\sum\limits_{i \in V}\frac{1}{N}\Big(1 - \prod\limits_{u \in P_{i}^z}(1-x_u)\Big)$ 
     & \makecell{$h\left(g_z(\vc{x})\right)$\\
     where\\
     $h(s) = \log(1+s)$
     } 
     & \makecell{$\hat{h}^L(g_z(\mathbf{x}))$, \\
     where $\hat{h}^L$ is 
     Eq.~\eqref{eq: taylor_f_iL}}
     & $\frac{1}{(L+1) 2^{L+1}}$ \\
     \hline
     \thead{FL} 
     & \makecell{Complete weighted bipartite\\
                 graph $G = (V \cup V')$\\
                 weights $w_{i_\ell, z} \in [0, 1]^{N \times |z|}$
                 } 
     & $\sum\limits_{\ell=1}^{N}(w_{i_\ell, z}-w_{i_{\ell+1}, z})\left(1-\prod\limits_{k=1}^\ell(1-x_{i_k})\right)$ 
     & \makecell{$h\left(g_z(\vc{x})\right)$\\
     where\\
     $h(s) = \log(1+s)$
     } 
     & \makecell{$\hat{h}^L(g_z(\mathbf{x}))$, \\
     where $\hat{h}^L$ is 
     Eq.~\eqref{eq: taylor_f_iL}} 
     & $\frac{1}{(L+1) 2^{L+1}}$ \\
     \hline
     \thead{CN} 
     & \makecell{Graph $G = (V, E)$,\\
                 service rates $\mu \in \reals_+^{|z|}$, \\
                 requests $r \in \mathcal{R}$, $P_z$ path of $r$, \\
                 arrival rates $\lambda \in \reals_+^{|\mathcal{R}|}$\\
                 } 
     & $\frac{1}{\mu_z}\sum_{r \in \mathcal{R}:z\in p^r} \lambda^r \prod_{k'=1}^{k_{p^r}(v)}(1-x_{p_k^r, i^r})$ 
     & \makecell{$h(g_z(\mathbf{0})) - h(g_z(\mathbf{x}))$\\
     where \\
     $h(s) = 
     {s}/{(1 - s)}$} 
     & \makecell{$\hat{h}^L(g_z(\mathbf{x}))$, \\
     where $\hat{h}^L$ is 
     Eq.~\eqref{eq: f_iL_CN}} 
     & $\frac{\bar{s}^{L+1}}{1-\bar{s}}$ \\
     \hline
    \end{tabular}}
    \vspace*{-10pt}
\end{table*}

\subsection{Data Summarization (SM)\cite{lin2011class, mirzasoleiman2016fast,kazemi2019submodular}}
In data summarization, ground set $V$ is a set of tokens, representing, e.g., words or sentences in a document. A corpus of documents $V_z$ is presented to us sequentially, and the goal is to select a ``summary'' $S\subseteq V$ that is representative of $V_z$. The summary should be simultaneously (a) representative of the corpus, and (b) diverse. 

To be representative, the summary $S\subset V$ should contain tokens of high value, where the value of a token is document-dependent: 
 for  document $z \in V_z$,  token $i \in V$  has a value $r_{i, z}\in [0,1]$, where $\sum_i r_{i, z}=1$. An example of such a value is the term frequency, i.e., the number of times the token appears in the document, divided by the document's length (in tokens). To be diverse, the summary should contain tokens that cover different subjects. To that end, if tokens are partitioned in to subjects, represented by a partition $\{P_j\}_{j=1}^J$ of $V$, the objective is given by $f(\mathbf{x}) = \mathbb{E}_z(f_z(\mathbf{x}))$ where 
 $   f_z(\mathbf{x}) = \textstyle\sum_{j=1}^J h\left(\sum_{i \in V\cap P_j} r_{i, z}x_i\right),$
and $h(s)=\log (1+s)$ is a non-decreasing concave function. 
Intuitively, the concavity of $h$ suppresses the selection of similar tokens (corresponding to the same subject), even if they have high value, thereby promoting diversity.
 Functions $f_z$ (and, thereby, also $f$) are monotone and submodular, and we can construct polynomial approximators $\hat{f}^L_z$ for them as indicated in Table~\ref{table: problems} by replacing $h$ with its 
 $L^{\text{th}}$-order Taylor approximation around 1/2, given by:
\begin{equation} \label{eq: taylor_f_iL}
    \hat{h}^{L}(s) = \textstyle\sum_{\ell = 0}^L \frac{h^{(\ell)}(1/2)}{\ell!} (s - {1}/{2})^\ell.
\end{equation}
This is because the composition of polynomial $\hat{f}^L_z$ with  polynomial  $g_z$ in Table~\ref{table: problems} is again a polynomial. 
We show in \fullversion{\cite{ozcan2021submodular}}{App.~\ref{app: proof_bias_bound} \deleted{of the supplement}} that this estimator ensures that $f$ indeed satisfies Asm.~\ref{asmp:boundedPoly}. Moreover,  
the estimator bias \emph{decays exponentially} with degree $L$ (see Tab.~\ref{table: problems} and App.~\ref{app:customBias}), meaning that polynomial number of terms suffice to reduce the bias to a desired level. 
A partition matroid could be used with this objective to enforce that no more than $k_\ell$ sentences come from $\ell$-th user, etc. 
 \subsection{Influence Maximization (IM) \cite{kempe2003maximizing, chen2009efficient}} \label{sec: IM}
Given a directed graph $G = (V, E)$, we wish to maximize the expected fraction of nodes reached if we infect a set of nodes $S\subseteq V$ and the infection spreads via, e.g., the Independent Cascade (IC) model \cite{kempe2003maximizing}. Adding a concave utility to the fraction can enhance the value of nodes reached in  early stages.  
Formally, let $z$ can be a random simulation trace of the IC model, and $P_v^z\subseteq V$ is the set of nodes reachable from $v$ in a random simulation of the IC model. Then, the objective can be written as $f(\mathbf{x}) = \mathbb{E}_{z \sim P} [f_z(\mathbf{x})]$ where
    $f_z(\vc{x}) =\textstyle h\left(g_z(\vc{x}) \right),$ 
$h(s)=\log (1+s)$, and
    $g_z(\vc{x}) =\textstyle \sum_{v \in V}\frac{1}{N}\big(1 - \prod_{i \in P_v^z}(1-x_i)\big)$ is the number of infected nodes under seed set  $\mathbf{x}$.  
Since functions $g_z:[0,1]^N\to [0,1]$ are multilinear, monotone submodular and $h:[0,1]\to\reals$  is non-decreasing and concave, 
$f$ satisfies Asm.~\ref{asm:monSub} \cite{ozcan2021submodular}. 
Again, we can construct $\hat{f}^L$ by replacing $h$ by $\hat{h}^L$, given by Eq.~\eqref{eq: taylor_f_iL}.
This again ensures that $f$ indeed satisfies Asm.~\ref{asmp:boundedPoly}, and 
 the estimator bias again decays exponentially 
(see Tab.~\ref{table: problems}  and \fullversion{\cite{ozcan2021submodular}}{App.~\ref{app:customBias}}). Partition matroid constraints could be used in this setting to bound the number of seeds from some group (e.g., males/females, people in a zip code, etc.).

\subsection{Facility Location (FL)\cite{mokhtari2018conditional}}
Given a weighted bipartite graph $G = (V \cup V_z)$ and weights $w_{i, z} \in [0, 1]$, $ i \in V$, $ z \in V_z$, we wish to maximize:
\begin{equation}\label{eq:originalFL}
    f(S) = \textstyle\mathbb{E}_{z \sim P} \left[h(\max_{i \in S} w_{i, z})\right], 
\end{equation}
where $h(s)=\log(1+s)$.
Intuitively, $V$ and $V'$ represent facilities and customers respectively and $w_{v, v'}$ is the utility of facility $v$ for customer $v'$. The goal is to select a subset of facility locations $S\subset{V}$ to maximize the total utility, assuming every customer chooses the facility with the highest utility in the selection $S$; again, adding the concave function $h$ adds diversity, favoring the satisfaction of customers that are not already covered. This too becomes a coverage problem 
by observing that \cite{karimi2017stochastic}:
    $\max_{i \in S} w_{i, z}  = \sum\limits_{\ell=1}^{n}(w_{i_\ell, z}-w_{i_{\ell+1}, z})\big(1-\prod\limits_{k=1}^\ell(1-x_{i_k})\big),$
where,
 for a given $z \in V_z$, weights have been pre-sorted in a descending order as $w_{i_1,z} \geq \ldots \geq w_{i_n,z}$. 
and 
$w_{i_{n+1},j} \triangleq 0$. 
In a manner similar to Sec~\ref{sec: IM}, we can show that this function again satisfies Asm.~\ref{asm:monSub} and~\ref{asmp:boundedPoly}, using again the $L^{\text{th}}$-order Taylor approximation of $h$, given by Eq.~\eqref{eq: taylor_f_iL}; this will again lead to a bias that decays exponentially (see Tab.~\ref{table: problems} and App.~\ref{app:customBias}). 
We can again optimize such an objective over arbitrary matroids, which can enforce, e.g., that no more than $k$ facilities are selected from a geographic area or some other partition of $V$.

\section{Experiments}
  We evaluate Alg.~\ref{alg: SCG}
  , 
  with sampling and polynomial estimators over two well-known problem instances (influence maximization and facility location
  ) with real and synthetic 
  datasets. We summarize these setups in Tab.~\ref{tab:datasets}. For a more detailed overview of the datasets and experiment parameters, please refer to App.~\ref{app:exps}\deleted{of the supplement}. Our code \replaced{is publicly accessible}{will be public once the submission is reviewed}.\footnote{\url{https://github.com/neu-spiral/StochSubMax}}
  
\begin{wraptable}{r}{6cm}
\vspace*{-25pt}
\begin{center}
    \begin{tabular}{|c|c|ccc|cc|}
    \hline
    \thead{instance} & \thead{dataset} & \thead{$|z|$} & \thead{$|S|$} & \thead{$|E|$} & \thead{m} & \thead{k} \\
    \hline
    \thead{IM} & \texttt{SBPL} & 20 & 400 & 914 & 4 & 1 \\
    \thead{IM} & \texttt{ZKC} & 20 & 34 & 78 & 2 & 3 \\
    \thead{FL} & \texttt{MovieLens} & 4000 & 6041 & 256 & 10 & 2 \\
    \hline
    \end{tabular}
\caption{{Datasets and Experiment Parameters.}}\label{tab:datasets}\end{center}
\vspace*{-25pt}
\end{wraptable}
 
\noindent\textbf{Algorithms.} We compare the performance of different estimators. These estimators are: (a) sampling estimator (SAMP) with $N = 1, 10, 20, 100$ 
and (b) polynomial estimator (POLY) with $L = 1, 2
$. 


\noindent\textbf{Metrics.} We evaluate the performance of the estimators with their clock running time and via 
the maximum result ($\max f(\mathbf{y})$) obtained using the best available estimator for a given setting.

\begin{figure}[t]
\centering
\subfigure[\texttt{SyntheticBipartitePowerLaw}]{
\begin{minipage}{0.31\linewidth}
\centering
\includegraphics[width=1\linewidth]{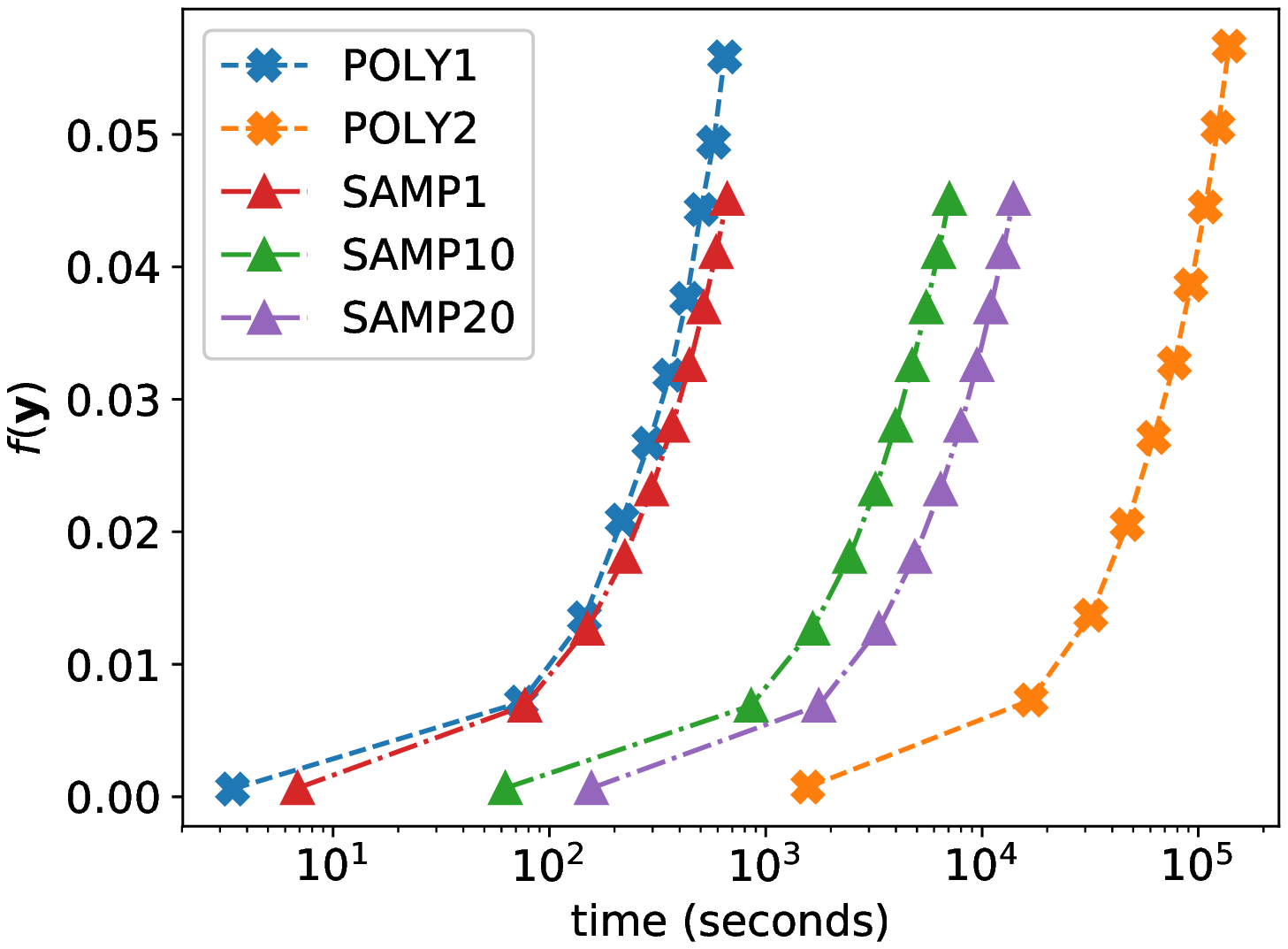}
\centering
\label{fig:SBPL_loglog}\vspace*{-10pt}
\end{minipage}
}
\subfigure[\texttt{ZKC}]{
\begin{minipage}{0.31\linewidth}
\centering
\includegraphics[width=1\linewidth]{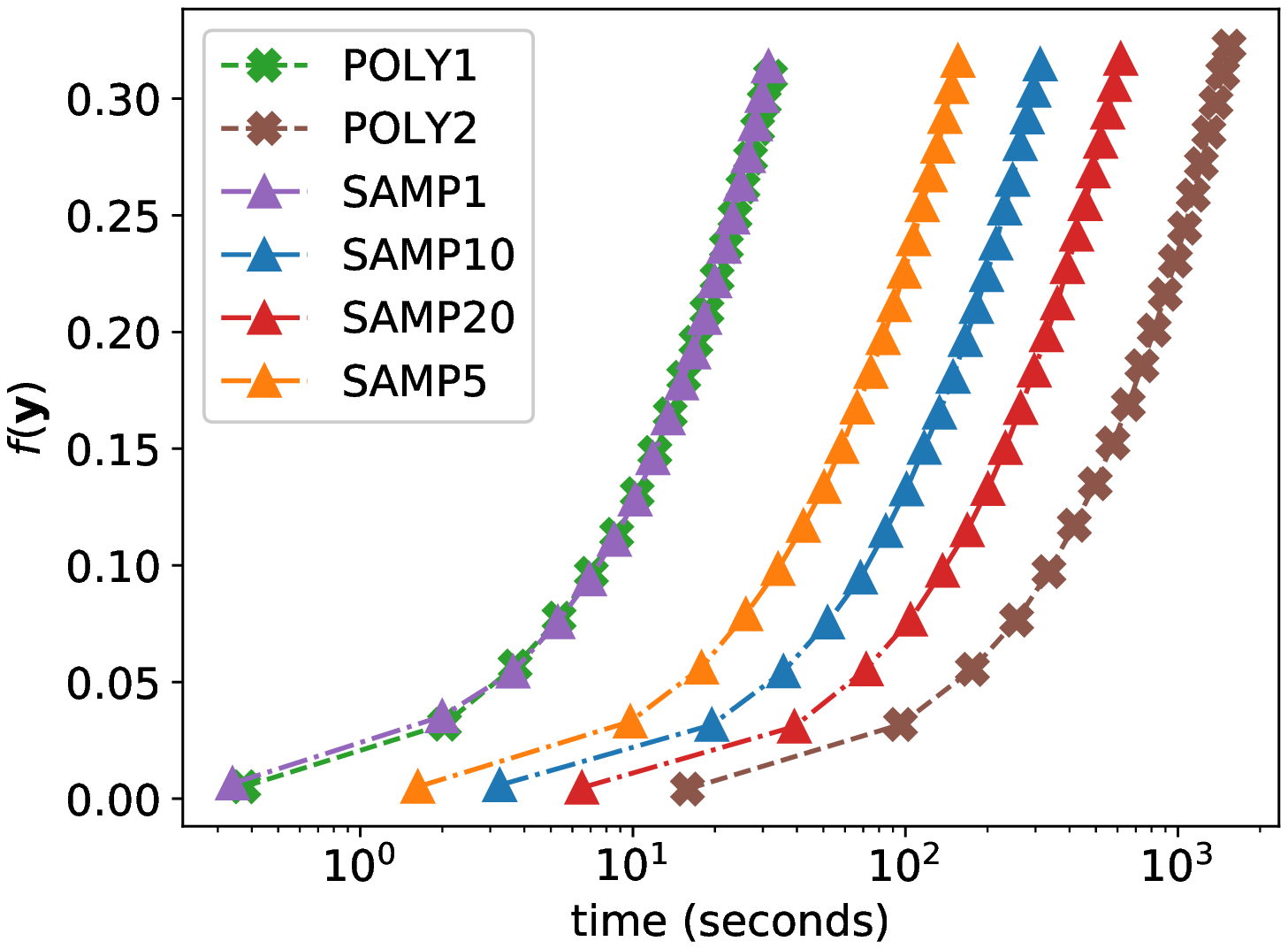}
\label{fig:ZKC_loglog}\vspace*{-10pt}
\end{minipage}
}
\subfigure[\texttt{MovieLens}]{
\begin{minipage}{0.31\linewidth}
\centering
\includegraphics[width=1\linewidth]{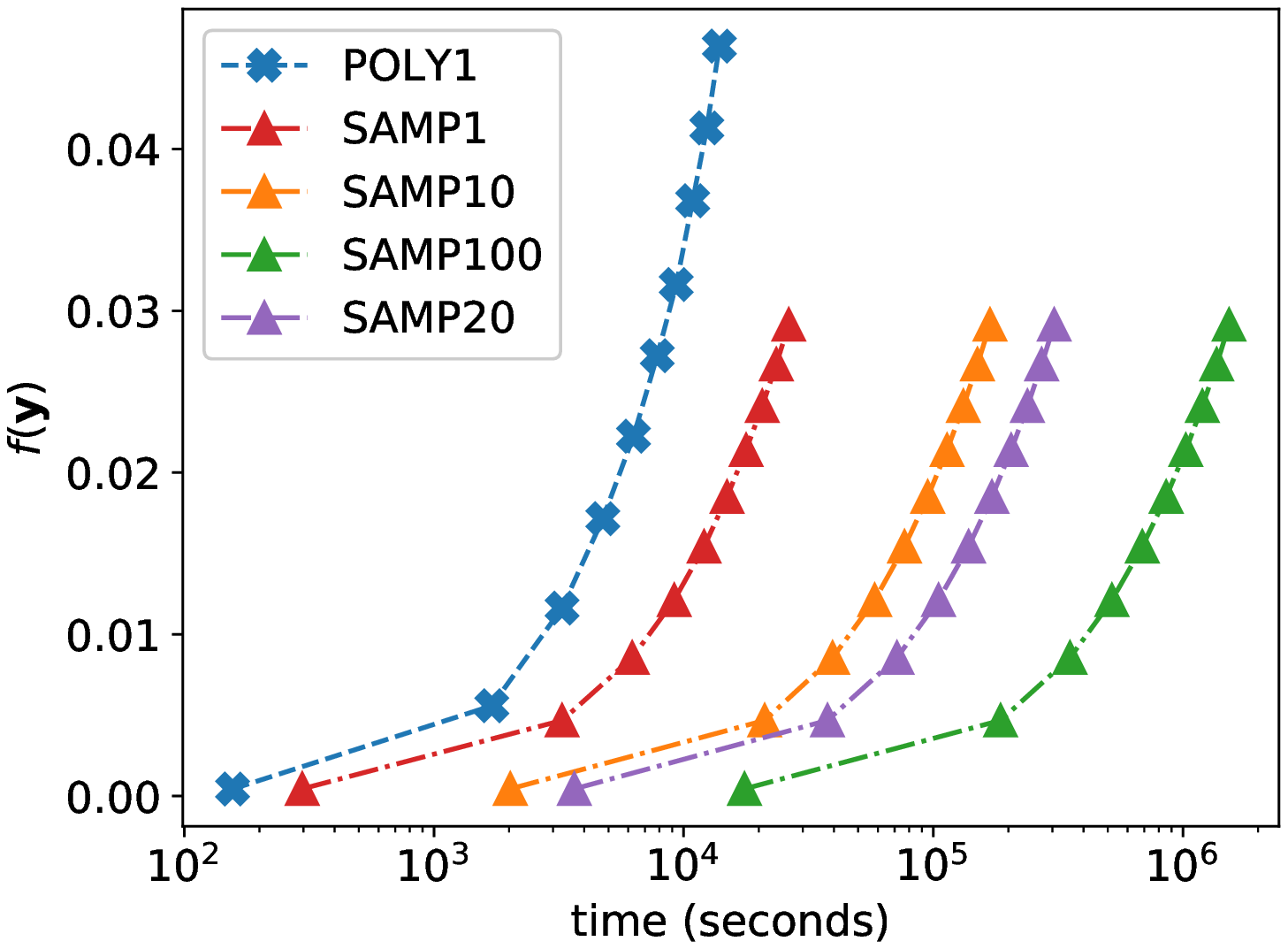}
\centering
\label{fig:MovieLens_loglog}
\end{minipage}
}
\vspace*{-10pt}
\caption{Trajectory of the FW algorithm. Utility of the function at the current $\vc{y}$ as a function of time is marked for every 
iteration.} 
 \vspace*{-13pt}
\label{fig:CGiters}
\end{figure}

\noindent\textbf{Results.} The trajectory of the utility obtained at each iteration of the stochastic continuous greey algorithm $f(\mathbf{y})$ is plotted as a function of time in Fig.~\ref{fig:CGiters}. 
In Fig.~\ref{fig:SBPL_loglog}, we observe that polynomial estimators outperforms sampling estimators in terms of utility. Moreover, POLY1 runs $10$ times faster than SAMP20 and runs in comparable time to SAMP1. In Fig.~\ref{fig:ZKC_loglog}, POLY2 outperforms all estimators whereas POLY1 slightly underperforms. Finally, in Fig.~\ref{fig:MovieLens_loglog} we observe that POLY1 consistently outperforms sampling estimators.

\begin{figure}[t]
\centering
\subfigure[\texttt{SBPL}]{
\begin{minipage}{0.30\linewidth}
\centering
\includegraphics[width=1\linewidth]{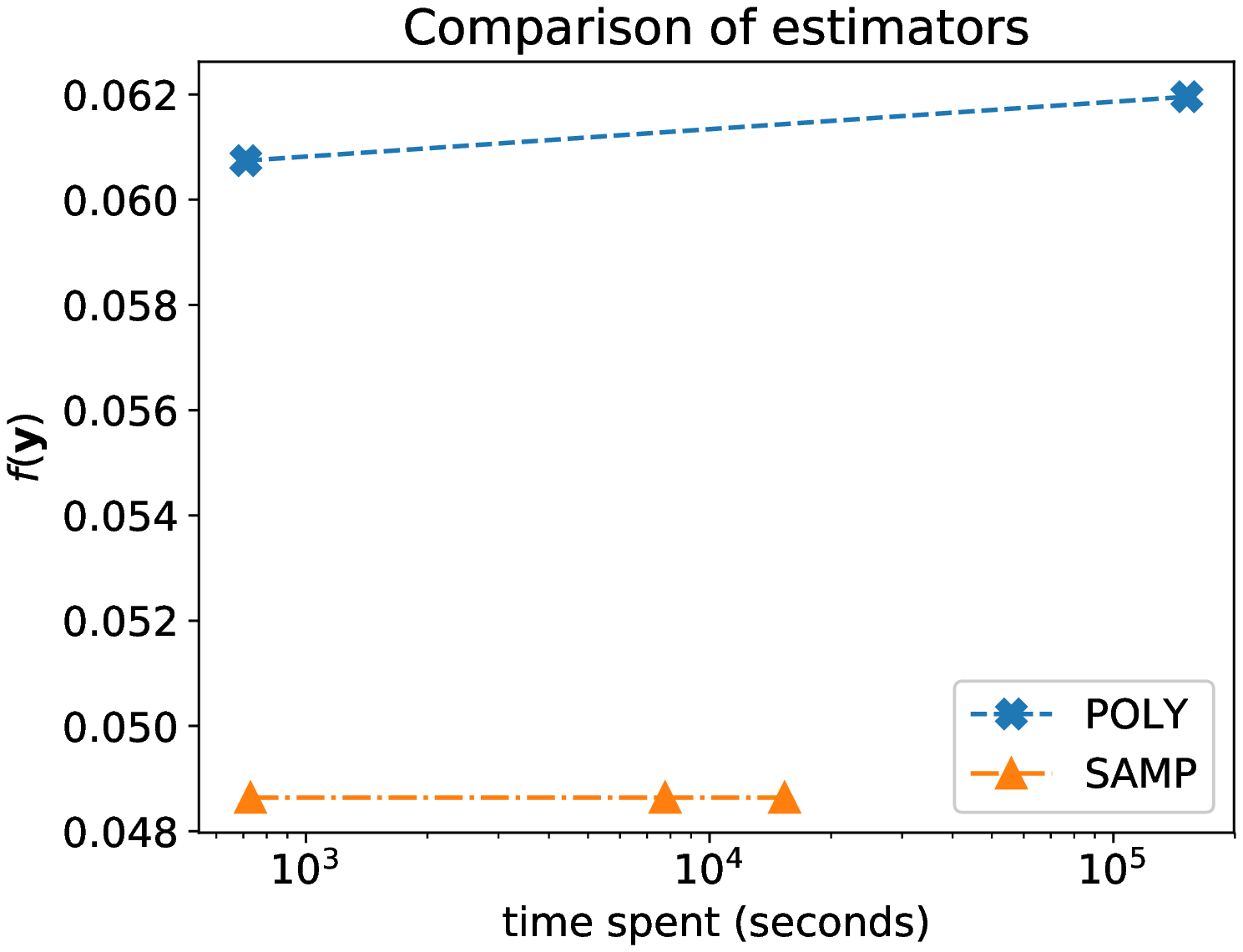}
\label{fig:SBPL_paretolog}\vspace*{-12pt}
\end{minipage}
}
\subfigure[\texttt{ZKC}]{
\begin{minipage}{0.30\linewidth}
\centering
\includegraphics[width=1\linewidth]{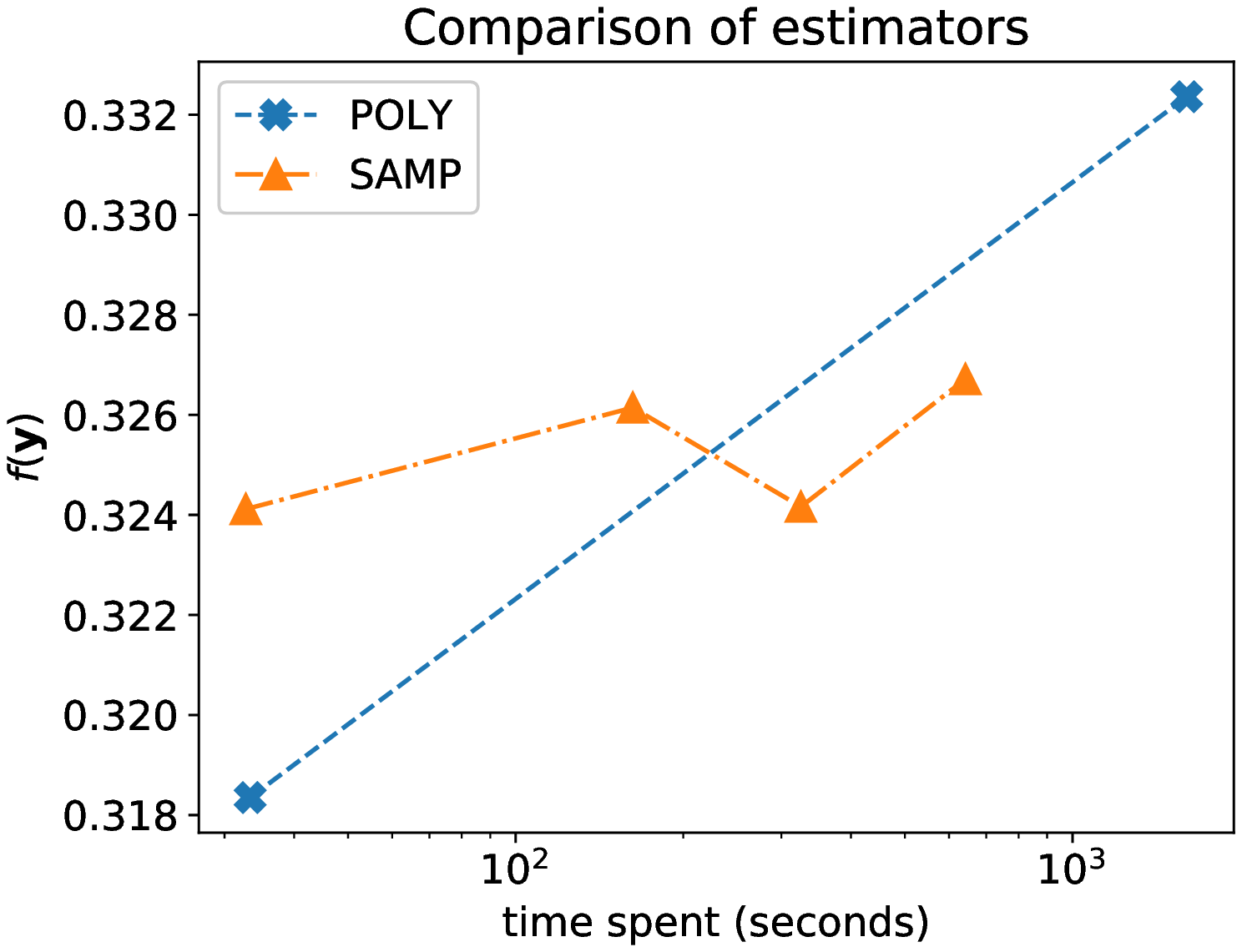}
\label{fig:ZKC_paretolog}\vspace*{-12pt}
\end{minipage}
}
\subfigure[\texttt{MovieLens}]{
\begin{minipage}{0.30\linewidth}
\centering
\includegraphics[width=1\linewidth]{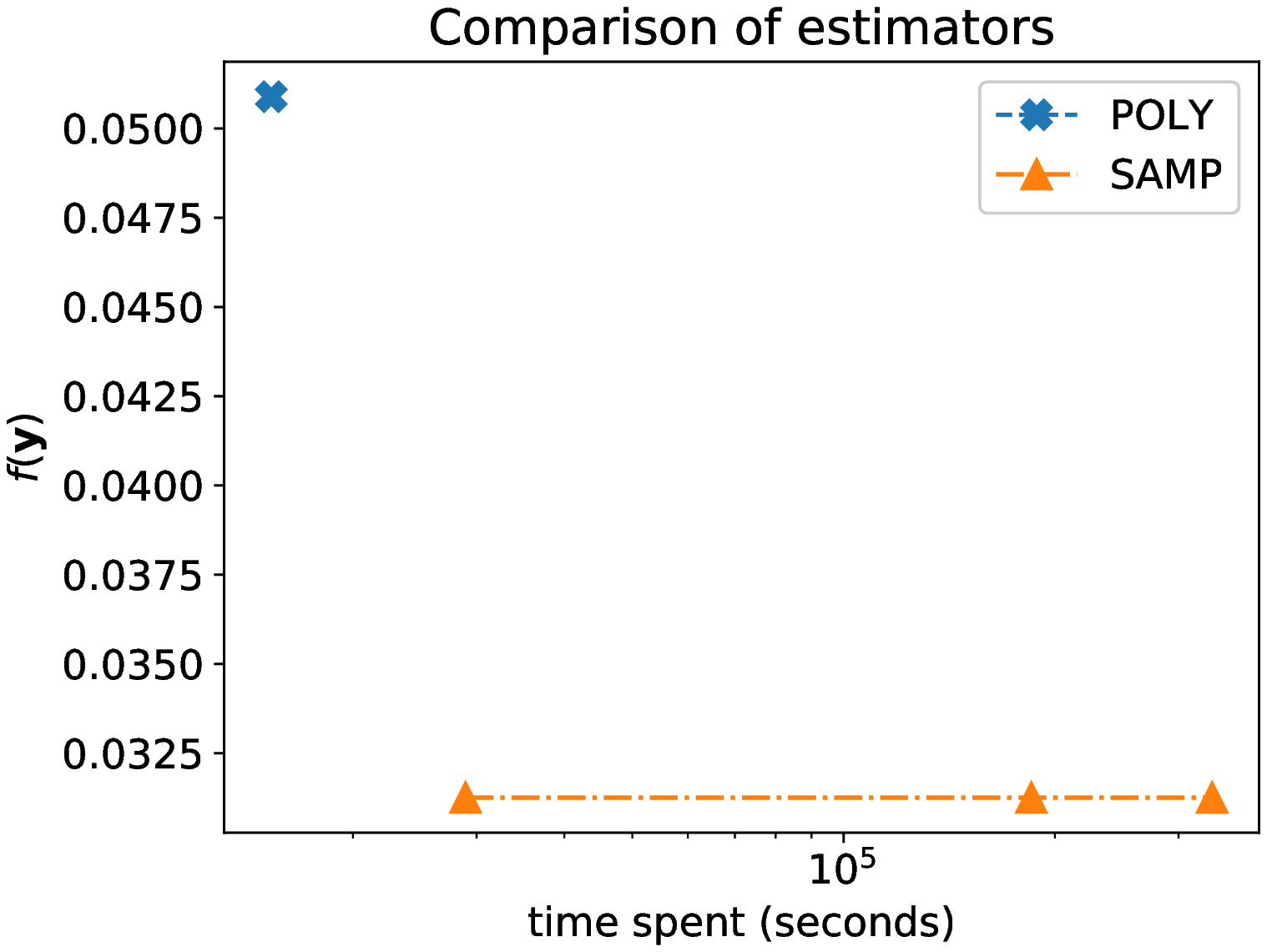}
\label{fig:MovieLens_paretolog}\vspace*{-12pt}
\end{minipage}
}
\vspace*{-10pt}
\caption{Comparison of different estimators on different problems. Blue lines represent the performance of the POLY estimators and the marked points correspond to POLY1 and POLY2 
respectively. Orange lines represent the performance of the SAMP estimators and the marked points correspond to SAMP1, SAMP10, SAMP20, SAMP100 respectively.}
\label{fig:final_estimates}
\end{figure}

The final outcomes of the objective functions of the estimators are reported as a function of time in Fig.~\ref{fig:final_estimates}. In Fig.~\ref{fig:SBPL_paretolog} and~\ref{fig:ZKC_paretolog}, POLY2 outperforms other estimators in terms of utility. Again in Fig.~\ref{fig:SBPL_paretolog}, POLY1 outperforms sampling estimators in terms of utility and runs in comparable time to SAMP1 while in Fig.~\ref{fig:MovieLens_paretolog}, POLY1 outperforms sampling estimators both in terms of time and utility. 
Ideally, we would expect the performance of the estimators to improve as the degree of the polynomial or the number of samples increase. The examples where this is not always the case can be explained by the stochastic nature of the problem.

\section{Conclusions}
We have shown that polynomial estimators can improve existing stochastic submodular maximization methods by eliminating one of the two sources of randomness, particularly the one that stems from sampling. Investigating methodical ways to construct such polynomials can expand the applications of the proposed estimator appearing in this paper. Online versions of stochastic submodular optimization, where performance is characterized in terms of (approximate) regret, are also a possible future research direction.



\bibliographystyle{splncs04}
\bibliography{08_ref}

\begin{thebibliography}{10}
\providecommand{\url}[1]{\texttt{#1}}
\providecommand{\urlprefix}{URL }
\providecommand{\doi}[1]{https://doi.org/#1}

\bibitem{ageev2004pipage}
Ageev, A.A., Sviridenko, M.I.: Pipage rounding: A new method of constructing
  algorithms with proven performance guarantee. Journal of Combinatorial
  Optimization  (2004)

\bibitem{asadpour2008stochastic}
Asadpour, A., Nazerzadeh, H., Saberi, A.: Stochastic submodular maximization.
  In: International Workshop on Internet and Network Economics (2008)

\bibitem{bateni2019categorical}
Bateni, M., Chen, L., Esfandiari, H., Fu, T., Mirrokni, V., Rostamizadeh, A.:
  Categorical feature compression via submodular optimization. In: ICML (2019)

\bibitem{broida1989comprehensive}
Broida, J., Williamson, S.: A Comprehensive Introduction to Linear Algebra.
  Advanced book program, Addison-Wesley (1989)

\bibitem{calinescu2011maximizing}
Calinescu, G., Chekuri, C., Pal, M., Vondr{\'a}k, J.: Maximizing a monotone
  submodular function subject to a matroid constraint. SICOMP  (2011)

\bibitem{chekuri2010dependent}
Chekuri, C., Vondr{\'a}k, J., Zenklusen, R.: Dependent randomized rounding via
  exchange properties of combinatorial structures. In: FoCS. IEEE (2010)

\bibitem{chen2018online}
Chen, L., Hassani, H., Karbasi, A.: Online continuous submodular maximization.
  In: AISTATS (2018)

\bibitem{chen2009efficient}
Chen, W., Wang, Y., Yang, S.: Efficient influence maximization in social
  networks. In: KDD (2009)

\bibitem{goyal2011data}
Goyal, A., Bonchi, F., Lakshmanan, L.V.: A data-based approach to social
  influence maximization. VLDB Endowment  (2011)

\bibitem{movielens}
Harper, F.M., Konstan, J.A.: The movielens datasets: History and context. TiiS
  (2015)

\bibitem{hassani2017gradient}
Hassani, H., Soltanolkotabi, M., Karbasi, A.: Gradient methods for submodular
  maximization. NeurIPS  (2017)

\bibitem{karimi2017stochastic}
Karimi, M.R., Lucic, M., Hassani, H., Krause, A.: Stochastic submodular
  maximization: the case of coverage functions. In: NeurIPS17 (2017)

\bibitem{kazemi2019submodular}
Kazemi, E., Mitrovic, M., Zadimoghaddam, M., Lattanzi, S., Karbasi, A.:
  Submodular streaming in all its glory: Tight approximation, minimum memory
  and low adaptive complexity. In: ICML (2019)

\bibitem{kempe2003maximizing}
Kempe, D., Kleinberg, J., Tardos, {\'E}.: Maximizing the spread of influence
  through a social network. In: KDD (2003)

\bibitem{lei2019discrete}
Lei, Q., Wu, L., Chen, P.Y., Dimakis, A., Dhillon, I.S., Witbrock, M.J.:
  Discrete adversarial attacks and submodular optimization with applications to
  text classification. MLSys  (2019)

\bibitem{lin2011class}
Lin, H., Bilmes, J.: A class of submodular functions for document
  summarization. In: ACL (2011)

\bibitem{mahdian2020kelly}
Mahdian, M., Moharrer, A., Ioannidis, S., Yeh, E.: Kelly cache networks.
  IEEE/ACM Transactions on Networking  (2020)

\bibitem{mirzasoleiman2016fast}
Mirzasoleiman, B., Badanidiyuru, A., Karbasi, A.: Fast constrained submodular
  maximization: Personalized data summarization. In: ICML (2016)

\bibitem{mokhtari2018conditional}
Mokhtari, A., Hassani, H., Karbasi, A.: Conditional gradient method for
  stochastic submodular maximization: Closing the gap. In: AISTATS (2018)

\bibitem{mokhtari2020stochastic}
Mokhtari, A., Hassani, H., Karbasi, A.: Stochastic conditional gradient
  methods: From convex minimization to submodular maximization. JMLR  (2020)

\bibitem{nemhauser1978analysis}
Nemhauser, G.L., Wolsey, L.A., Fisher, M.L.: An analysis of approximations for
  maximizing submodular set functions—i. Mathematical programming  (1978)

\bibitem{ozcan2021submodular}
{\"O}zcan, G., Moharrer, A., Ioannidis, S.: Submodular maximization via taylor
  series approximation. In: SDM (2021)

\bibitem{rossetti2017ndlib}
Rossetti, G., Milli, L., Rinzivillo, S., Sirbu, A., Pedreschi, D., Giannotti,
  F.: Ndlib: Studying network diffusion dynamics. In: DSAA. IEEE (2017)

\bibitem{susubmodular}
Su, F., Zhu, Y., Wu, O., Deng, Y.: Submodular meta data compiling for meta
  optimization

\bibitem{vondrak2008optimal}
Vondr{\'a}k, J.: Optimal approximation for the submodular welfare problem in
  the value oracle model. In: STOC (2008)

\bibitem{zachary1977information}
Zachary, W.W.: An information flow model for conflict and fission in small
  groups. Journal of anthropological research  (1977)

\bibitem{zhang2022stochastic}
Zhang, Q., Deng, Z., Chen, Z., Hu, H., Yang, Y.: Stochastic continuous
  submodular maximization: Boosting via non-oblivious function. In: ICML (2022)

\end{thebibliography}

\newpage
\section*{Appendix}
\addcontentsline{toc}{section}{Appendices}
\renewcommand{\thesubsection}{\Alph{subsection}}
\subsection{Proof of Lemma~\ref{lem:gradientBias}} \label{app:proof_gradBiasLemma}
    We start by showing that the norm of the residual error vector of the estimator converges to $0$. 
    Recall that, by Asm.~\ref{asmp:boundedPoly} the residual error of the polynomial estimation $\hat{f}_{z}^L(\mathbf{x})$ is bounded by $\varepsilon (L)$. Thus, for functions $f_z:\{0,1\}^n\rightarrow\reals_+$ satisfying Asm.~\ref{asmp:boundedPoly}, we have that for all $\mathbf{y} \in [0, 1]^n$,
    \begin{align}\label{eq:epsilon_zL}
        \lvert f_z(\mathbf{x}) - \hat{f}_z^L(\mathbf{x})\rvert\leq \varepsilon_z (L).
    \end{align}
    Since $\lim_{L\to\infty}\varepsilon_z (L)=0$ for all $\mathbf{y} \in [0,1]^n$, we get that
    \begin{align}
        \lim_{L\to\infty}\lvert f_z(\vc{y})-\hat{f_z^L}(\vc{y})\rvert\leq\lim_{L\to\infty}\varepsilon_z(L)=0.\label{eq:rl}
    \end{align}
    Moreover,
    \begin{align*}
         \left|\frac{\partial G_z(\vc{y})}{\partial y_i}-\frac{\widehat{\partial G_z^L}(\vc{y})}{\partial y_i}\right|&=\big|\mathbb{E}_{\vc{x}\sim\vc{y}}[f_z([\vc{x}]_{+i})]-\mathbb{E}_{\vc{x}\sim\vc{y}}[f_z([\vc{x}]_{-i})]\displaybreak[0] - \mathbb{E}_{\vc{x}\sim\vc{y}}[\hat{f_z^L}([\vc{x}]_{+i})] + \mathbb{E}_{\vc{x}\sim\vc{y}}[\hat{f_z^L}([\vc{x}]_{-i})]\big|\\
        &\leq \mathbb{E}_{\vc{x}\sim\vc{y}}[|f_z([\vc{x}]_{+i})-\hat{f_z^L}(\vc{[\vc{x}]_{+i}})|]\displaybreak[0] + \mathbb{E}_{\vc{x}\sim\vc{y}}[|f_z([\vc{x}]_{-i}) - \hat{f_z^L}([\vc{x}]_{-i})|] \\
        &\stackrel{\mbox{\tiny{(\ref{eq:epsilon_zL})}}}{\leq} \mathbb{E}_{\vc{x}\sim\vc{y}}[\varepsilon_z (L))]+\mathbb{E}_{\vc{x}\sim\vc{y}}[\varepsilon_z (L))] = 2\varepsilon_z(L).
    \end{align*}
    Hence, we conclude that
   $     \big\|\nabla G_z(\vc{y}) - \widehat{\nabla G_z^L}(\vc{y})\big\|_2 \leq 2\sqrt{n}\varepsilon_z(L).$
\hspace{\stretch{1}} \qed

\subsection{Proof of Theorem~\ref{thm:main}} \label{app: proofMainThm}
\begin{proof}
    \begin{lemma} \label{lem:bounded_EGy}
        Consider Stochastic Continuous Greedy (SCG) outlined in Algorithm~\ref{alg: SCG}, and recall the definitions of the function $G$, the rank $r$, and $f_{\max} = \max_{i \in \{1, \ldots, n\}} f(\{i\})$. If Assumption~\ref{asm:boundedNorm} is satisfied, then for $t = 0, \ldots, T$ and for $j = 1, \ldots, n$ we have
            \begin{equation}
                \begin{split}
                \mathbb{E}\left[G(\mathbf{y}_{t+1})\right] &\geq \mathbb{E}\left[G(\mathbf{y}_t)\right] + \frac{1}{T} \mathbb{E}\left[G(\mathbf{y}^*) - G(\mathbf{y}_t)\right] - \frac{f_{\max}r D^2}{2T^2}\\
                &\quad - \frac{1}{2T}\left(4\beta_t D^2 - \frac{\mathbb{E}\left[\|\nabla G(\mathbf{y}_t) - \mathbf{d}_t\|^2\right]}{\beta_t} \right).
            \end{split}
            \end{equation}
    \end{lemma}

\begin{proof}
    According to the Taylor's expansion of the function $G$ near the point $\mathbf{y}_t$ we can write
        \begin{equation*}
            \begin{split}
                G(\mathbf{y}_{t+1}) &= G(\mathbf{y}_{t}) + \langle \nabla G(\mathbf{y}_{t}), \mathbf{y}_{t+1} - \mathbf{y}_{t} \rangle + \frac{1}{2} \langle \mathbf{y}_{t+1} - \mathbf{y}_{t}, \mathbf{H}(\tilde{\mathbf{y}}_t)(\mathbf{y}_{t+1} - \mathbf{y}_{t}) \rangle,\\
                &= G(\mathbf{y}_{t}) + \frac{1}{T}\langle \nabla G(\mathbf{y}_{t}), \mathbf{v}_{t} \rangle + \frac{1}{2T^2} \langle \mathbf{v}_{t}, \mathbf{H}(\tilde{\mathbf{y}}_t)\mathbf{v}_{t} \rangle
            \end{split}
        \end{equation*}
    where $\tilde{\mathbf{y}}_t$ is a convex combination of $\mathbf{y}_t$ and $\mathbf{y}_t + \frac{1}{T}\mathbf{v}_{t}$ and $\mathbf{H}(\tilde{\mathbf{y}}_t) = \nabla^2 G(\tilde{\mathbf{y}}_t)$. Note here that the elements of the matrix $\mathbf{H}(\tilde{\mathbf{y}}_t)$ are less than the maximum marginal value (i.e. 
    $\max_{i, j} |H_{i, j}(\tilde{\mathbf{y}}_t)| \leq \max_{i \in \{1, \ldots, n\}} f(\{i\}) = f_{\max}$). Therefore, we can lower bound $H_{ij}$ by $-f_{\max}$.
        \begin{equation*}
            \begin{split}
                \langle \mathbf{v}_{t}, \mathbf{H}(\tilde{\mathbf{y}}_t)\mathbf{v}_{t} \rangle &= \sum_{j=1}^n \sum_{i=1}^n v_{i, t} v_{j, t} H_{ij}(\tilde{\mathbf{y}}_t)\\
                &\geq -f_{\max} \sum_{j=1}^n \sum_{i=1}^n v_{i, t} v_{j, t} \\
                &= -f_{\max} \left(\sum_{i=1}^n v_{i, t}\right)^2 = -f_{\max} r \|\mathbf{v}_t\|^2,
            \end{split}
        \end{equation*}
    where the last equality is because $\mathbf{v}_t$ is a vector with $r$ ones and $n-r$ zeros. Replacing $\langle \mathbf{v}_{t}, \mathbf{H}(\tilde{\mathbf{y}}_t)\mathbf{v}_{t} \rangle$ by its lower bound $-f_{\max} r \|\mathbf{v}_t\|^2$ to obtain
        \begin{equation}\label{eq:maintheostep1}
            G(\mathbf{y}_{t+1}) \geq G(\mathbf{y}_{t}) + \frac{1}{T}\langle\nabla G(\mathbf{y}_t), \mathbf{v}_t\rangle - \frac{f_{\max}r}{2T^2} \|\mathbf{v}_t\|^2.
        \end{equation}
     
    Let $\mathbf{y}^*$ be the global maximizer within the constraint set $\mathcal{C}$. Since $\mathbf{v}_t$ is in the set $\mathcal{C}$, it follows from Assumption~\ref{asm:boundedNorm} that the norm $\|\mathbf{v}_t\|^2$ is bounded above by $D^2$. Apply this substitution and add and subtract the inner product $\frac{1}{T} \langle \mathbf{v}_t, \mathbf{d}_t \rangle$ to the right hand side of (\ref{eq:maintheostep1}) to obtain
        \begin{equation}
            \begin{split}
                G(\mathbf{y}_{t+1}) &\geq G(\mathbf{y}_t) + \frac{1}{T} \langle \mathbf{v}_t, \mathbf{d}_t \rangle + \frac{1}{T} \langle \mathbf{v}_t, \nabla G(\mathbf{y}_t) - \mathbf{d}_t \rangle - \frac{f_{\max}r D^2}{2T^2} \\
                &\geq G(\mathbf{y}_t) + \frac{1}{T} \langle \mathbf{y}^*, \mathbf{d}_t \rangle + \frac{1}{T} \langle \mathbf{v}_t, \nabla G(\mathbf{y}_t) - \mathbf{d}_t \rangle - \frac{f_{\max}r D^2}{2T^2}.\label{eq:maintheostep2}
            \end{split}
        \end{equation}
    Note that the second inequality in (\ref{eq:maintheostep2}) holds since $\mathbf{v}_t \in \arg\max_{\mathbf{v} \in \mathcal{C}} \{\mathbf{d}_t^T \mathbf{v}\}$, we can write
        \begin{equation}
            \langle \mathbf{y}^*, \mathbf{d}_t \rangle \leq \max_{\mathbf{v} \in \mathcal{C}} \{\langle \mathbf{v}, \mathbf{d}_t \rangle \} = \langle \mathbf{v}_t, \mathbf{d}_t \rangle.
        \end{equation}
    Now add and subtract the inner product $\langle \mathbf{y}^*, \nabla G(\mathbf{x}_t) - \mathbf{d}_t \rangle / T$ to the right hand side of (\ref{eq:maintheostep2}) to get
        \begin{equation}
            \begin{split}
                G(\mathbf{y}_{t+1}) &\geq G(\mathbf{y}_t) + \frac{1}{T} \langle \mathbf{y}^*, \nabla G(\mathbf{y}_t) \rangle + \frac{1}{T} \langle \mathbf{v}_t - \mathbf{y}^*, \nabla G(\mathbf{y}_t) - \mathbf{d}_t \rangle - \frac{f_{\max}r D^2}{2T^2}.
            \end{split} \label{eq:maintheostep3}
        \end{equation}
    We further have $\langle \mathbf{y}^*, \nabla G(\mathbf{y}_t) - \mathbf{d}_t \rangle \geq G(\mathbf{y}^*) - G(\mathbf{y}_t);$ this follows from monotonicity of $G$ as well as concavity of $G$ along positive directions; see, e.g., \cite{calinescu2011maximizing}. Moreover, by Young's inequality we can show that the inner product $\langle \mathbf{v}_t - \mathbf{y}^*, \nabla G(\mathbf{y}_t) - \mathbf{d}_t \rangle$ is lower bounded by
        \begin{equation}
            \begin{split}
                \langle \mathbf{v}_t - \mathbf{y}^*, \nabla G(\mathbf{x}_t) - \mathbf{d}_t \rangle \geq &
                -\frac{\beta_t}{2} \|\mathbf{v}_t - \mathbf{y}^*\|^2 - \frac{1}{2\beta_t} \|\nabla G(\mathbf{y}_t) - \mathbf{d}_t\|^2,
            \end{split}
        \end{equation}
    for any $\beta_t > 0.$ By applying these substitutions into (\ref{eq:maintheostep3}) we obtain
        \begin{equation}
            \begin{split}
                G(\mathbf{y}_{t+1}) &\geq G(\mathbf{y}_t) + \frac{1}{T} (G(\mathbf{y}^*) - G(\mathbf{y}_t)) - \frac{f_{\max}r D^2}{2T^2} - \frac{1}{2T}\left(\beta_t \|\mathbf{v}_t - \mathbf{y}^*\|^2 - \frac{\|\nabla G(\mathbf{y}_t) - \mathbf{d}_t\|^2}{\beta_t} \right)  .\label{eq:maintheostep4}
            \end{split}
        \end{equation}
    Replace $\|\mathbf{v}_t - \mathbf{y}^*\|^2$ by its upper bound $4D^2$ and compute the expected value of (\ref{eq:maintheostep4}) to write
        \begin{equation}
            \begin{split}
                \mathbb{E}\left[G(\mathbf{y}_{t+1})\right] &\geq \mathbb{E}\left[G(\mathbf{y}_t)\right] + \frac{1}{T} \mathbb{E}\left[G(\mathbf{y}^*) - G(\mathbf{y}_t)\right] - \frac{f_{\max}r D^2}{2T^2}\\
                &\quad - \frac{1}{2T}\left(4\beta_t D^2 - \frac{\mathbb{E}\left[\|\nabla G(\mathbf{y}_t) - \mathbf{d}_t\|^2\right]}{\beta_t} \right).
            \end{split}
        \end{equation}\label{eq:maintheostep5}
    \end{proof}
    
    We need to introduce Lemma~\ref{lem:bounded_dt} to provide a bound for $\mathbb{E}\left[\|\nabla G(\mathbf{y}_t) - \mathbf{d}_t\|^2\right]$ and in order to prove Lemma~\ref{lem:bounded_dt}, we need three new lemmas, Lemmas~\ref{lem:pointwiseLipschitz},~\ref{lem:bias_bound}, and \ref{lem:recursive}, respectively.
    
    \begin{lemma} \label{lem:pointwiseLipschitz}
        Consider Stochastic Continuous Greedy (SCG) outlined in Algorithm~\ref{alg: SCG} with iterates $\mathbf{y}_t$, and recall the definition of the multilinear extension function $G$ in (\ref{eq: multilinear}). If we define $r$ as the rank of the matroid $\mathcal{I}$ and $f_{\max} = \max_{i\in\{1, \ldots, n\}} f(i),$ then the following holds
        $$|\nabla_j G(\mathbf{y}_{t+1}) - \nabla_j G(\mathbf{y}_t)| \leq f_{\max} \sqrt{r} \|\mathbf{y}_{t+1} - \mathbf{y}_t\|,$$
        for $j = 1, \ldots, n.$
    \end{lemma}
    
    \begin{proof}
        Same as the proof of Lemma~11 in \cite{mokhtari2020stochastic}.
    \end{proof}
    
    \begin{lemma}\label{lem:bias_bound}
    The variance of the biased stochastic gradients $\widehat{\nabla G_{z_t}^L(\mathbf{y}_t)}$ is bounded above by $\left(\sigma_0 + 2\sqrt{n}\varepsilon (L)\right)^2$, i.e., for any vector $\mathbf{y} \in \mathcal{C}$ we can write
        \begin{equation}
            \begin{split}
            \mathbb{E}\left[\left\|\nabla G(\mathbf{y}) - \widehat{\nabla G_{z}^L(\mathbf{y})}\right\|^2 \right] &\leq (1 + \beta_0)\sigma_0^2 + \left(1 + \frac{1}{\beta_0}\right)2\sqrt{n}\varepsilon (L),
            \end{split}
        \end{equation}
    where the expectation is with respect to the randomness of $z \sim P$.
    \end{lemma}
    \begin{proof}    
    Adding and subtracting $\nabla G_{z_t}(\mathbf{y}_t)$ inside the norm, we obtain
        \begin{equation}
            \begin{split}
                \mathbb{E}\left[\left\|\nabla G(\mathbf{y}) - \widehat{\nabla G_{z_t}^L(\mathbf{y})}\right\|^2 \right]  &= \mathbb{E}\left[\left\|\nabla G(\mathbf{y}) - \nabla G_{z}(\mathbf{y}) + \nabla G_{z}(\mathbf{y}) -  \widehat{\nabla G_{z}^L(\mathbf{y})}\right\|^2 \right],\\
                &= \mathbb{E}\bigg[\left\|\nabla G(\mathbf{y}) - \nabla G_{z}(\mathbf{y})\right\|^2 \\
                &+ 2\left(\nabla G(\mathbf{y}) - \nabla G_{z}(\mathbf{y})\right)^T\left(\nabla G_{z}(\mathbf{y}) -  \widehat{\nabla G_{z}^L(\mathbf{y})}\right) \\
                &+ \left\|\nabla G_{z}(\mathbf{y}) -  \widehat{\nabla G_{z}^L(\mathbf{y})}\right\|^2 \bigg],\\
                &= \mathbb{E}\left[\left\|\nabla G(\mathbf{y}) - \nabla G_{z}(\mathbf{y})\right\|^2 \right] \\
                &+ 2\mathbb{E}\left[\left(\nabla G(\mathbf{y}) - \nabla G_{z}(\mathbf{y})\right)^T\left(\nabla G_{z}(\mathbf{y}) -  \widehat{\nabla G_{z}^L(\mathbf{y})}\right) \right] \\
                &+ \mathbb{E}\left[\left\|\nabla G_{z}(\mathbf{y}) -  \widehat{\nabla G_{z}^L(\mathbf{y})}\right\|^2 \right].
            \end{split}
        \end{equation}
        
    Using Young's inequality $\left(\langle \mathbf{a}, \mathbf{b}\rangle \leq \frac{1}{\beta}\frac{\|\mathbf{a}\|^2}{2} + \beta\frac{\|\mathbf{b}\|^2}{2}\right)$, also known as Peter-Paul inequality, to substitute the inner products with summations to obtain
    \begin{equation}
            \begin{split}
                \mathbb{E}\left[\left\|\nabla G(\mathbf{y}) - \widehat{\nabla G_{z_t}^L(\mathbf{y})}\right\|^2 \right]  &= \mathbb{E}\left[\left\|\nabla G(\mathbf{y}) - \nabla G_{z}(\mathbf{y}) + \nabla G_{z}(\mathbf{y}) -  \widehat{\nabla G_{z}^L(\mathbf{y})}\right\|^2 \right],\\
                &\leq \mathbb{E}\left[\left\|\nabla G(\mathbf{y}) - \nabla G_{z}(\mathbf{y})\right\|^2 \right] \\
                &\quad + \mathbb{E}\bigg[\beta_0\left\|\nabla G(\mathbf{y}) - \nabla G_{z}(\mathbf{y})\right\|^2 + \frac{1}{\beta_0}\left\|\nabla G_{z}(\mathbf{y}) -  \widehat{\nabla G_{z}^L(\mathbf{y})}\right\|^2 \bigg] \\
                &\quad + \mathbb{E}\left[\left\|\nabla G_{z}(\mathbf{y}) -  \widehat{\nabla G_{z}^L(\mathbf{y})}\right\|^2 \right],\\
                &\leq (1 + \beta_0)\mathbb{E}\left[\left\|\nabla G(\mathbf{y}) - \nabla G_{z}(\mathbf{y})\right\|^2 \right] \\
                &\quad + \left(1 + \frac{1}{\beta_0}\right)\mathbb{E}\left[\left\|\nabla G_{z}(\mathbf{y}) -  \widehat{\nabla G_{z}^L(\mathbf{y})}\right\|^2 \right].
            \end{split}
        \end{equation}
    
     Replacing $\mathbb{E}\left[\left\|\nabla G(\mathbf{y}) - \nabla G_{z}(\mathbf{y})\right\|^2 \right]$ by its upper bound $\sigma_0^2$ and using the result of Lemma~\ref{lem:gradientBias} to replace $\mathbb{E}\left[\left\|\nabla G_{z}(\mathbf{y}) - \widehat{\nabla G_{z}^L(\mathbf{y})}\right\|^2 \right]$ by its upper bound $2\sqrt{n}\varepsilon (L)$, we obtain
    \begin{equation}
            \begin{split}
                \mathbb{E}\left[\left\|\nabla G(\mathbf{y}) - \widehat{\nabla G_{z}^L(\mathbf{y})}\right\|^2 \right] &\leq (1 + \beta_0)\sigma_0^2 + \left(1 + \frac{1}{\beta_0}\right)2\sqrt{n}\varepsilon (L).
            \end{split}
        \end{equation} 
    \end{proof}

    \begin{lemma}\label{lem:recursive}
        (Directly from \cite{mokhtari2020stochastic}) Consider the scalars $b \geq 0$ and $c > 1.$ Let $\phi_t$ be a sequence of real numbers satisfying $$\phi_t \leq \left(1 - \frac{c}{(t+t_0)^{\alpha}}\right) \phi_{t-1} + \frac{b}{(t+t_0)^{2\alpha}},$$ for some $0 \leq \alpha \leq 1$ and $t_0 \geq 0.$ Then, the sequence $\phi_t$ converges to zero at the following rate $$\phi_t \leq \frac{Q}{(t+t_0+1)^{\alpha}},$$ where $Q = \max\{\phi_0 (t_0 + 1)^{\alpha}, b/(c-1)\}.$
    \end{lemma}
    \begin{proof}
    Proof of the lemma can be found in the Appendix~C of \cite{mokhtari2020stochastic}.
    \end{proof}
    
    \begin{lemma} \label{lem:bounded_dt}
        Consider Stochastic Continuous Greedy (SCG) outlined in Algorithm~\ref{alg: SCG}, and recall the definitions of the function $G$, the rank $r$, the upper bound $\sigma_0$ defined as in Thm.~\ref{thm:main} and $f_{\max} = \max_{i \in \{1, \ldots, n\}} f(\{i\})$. If Assumption~\ref{asm:boundedNorm} 
        is satisfied and $\rho_t = \frac{4}{(t+8)^{2/3}},$ then for $t = 0, \ldots, T$ and for $j = 1, \ldots, n$ we have
            \begin{equation}
                \mathbb{E}\left[\|\nabla G(\mathbf{y}_t) - \mathbf{d}_t\|^2\right] \leq \frac{Q}{(t+9)^{2/3}},
            \end{equation}
        where $Q = \max\{5\|\nabla G(\mathbf{y}_0 - \mathbf{d}_0)\|^2, 32\sigma_0^2 + 224\sqrt{n}\varepsilon (L) + 4f_{\max}^2rD^2\}.$
    \end{lemma}

    \begin{proof}
    Use the definition $\mathbf{d}_t = (1 - \rho_t) \mathbf{d}_{t-1} + \rho_t \nabla \widehat{G_{z_t}^L (\mathbf{y}_t)}$ to write $\|\nabla G(\mathbf{y}_t) - \mathbf{d}_t\|^2$ as
        \begin{equation}
            \begin{split}
                \|\nabla G(\mathbf{y}_t) - \mathbf{d}_t\|^2 = \|&\nabla G(\mathbf{y}_t) - (1 - \rho_t) \mathbf{d}_{t-1} - \rho_t \nabla \widehat{G_{z_t}^L (\mathbf{y}_t)}\|^2.\label{eq:dt_bound_step1}
            \end{split}
        \end{equation}
    Add and subtract the term $(1 - \rho_t) \nabla G(\mathbf{y}_{t-1})$ to the right hand side of (\ref{eq:dt_bound_step1}), regroup the terms and expand the squared term to obtain
    \begin{equation}
        \begin{split}
            \|\nabla G(\mathbf{y}_t) - \mathbf{d}_t\|^2 &= \|\nabla G(\mathbf{y}_t) - (1 - \rho_t) \nabla G(\mathbf{y}_{t-1}) + (1 - \rho_t) \nabla G(\mathbf{y}_{t-1}) \\
            &\quad - (1 - \rho_t) \mathbf{d}_{t-1} - \rho_t \nabla \widehat{G_{z_t}^L (\mathbf{y}_t)}\|^2\\
            &= \|\rho_t(\nabla G(\mathbf{y}_t) - \nabla \widehat{G_{z_t}^L (\mathbf{y}_t)}) + (1 - \rho_t) (\nabla G(\mathbf{y}_t) - \nabla G(\mathbf{y}_{t-1})) \\
            &\quad + (1 - \rho_t) (\nabla G(\mathbf{y}_{t-1})- \mathbf{d}_{t-1}) \|^2\\
            &= \rho_t^2 \|\nabla G(\mathbf{y}_t) - \nabla \widehat{G_{z_t}^L (\mathbf{y}_t)}\|^2
            + (1 - \rho_t)^2 \|\nabla G(\mathbf{y}_t) - \nabla G(\mathbf{y}_{t-1})\|^2 \\
            &\quad + (1 - \rho_t)^2 \|\nabla G(\mathbf{y}_{t-1})- \mathbf{d}_{t-1} \|^2 \\
            & \quad + 2\rho_t(1 - \rho_t)(\nabla G(\mathbf{y}_t) - \nabla \widehat{G_{z_t}^L (\mathbf{y}_t)})^T (\nabla G(\mathbf{y}_t) - \nabla G(\mathbf{y}_{t-1})) \\
            &\quad + 2\rho_t(1 - \rho_t)(\nabla G(\mathbf{y}_t) - \nabla \widehat{G_{z_t}^L (\mathbf{y}_t)})^T (\nabla G(\mathbf{y}_{t-1})- \mathbf{d}_{t-1})\\
            &\quad + 2(1 - \rho_t)^2(\nabla G(\mathbf{y}_t) - \nabla G(\mathbf{y}_{t-1})^T (\nabla G(\mathbf{y}_{t-1})- \mathbf{d}_{t-1}).
        \end{split} \label{eq:dt_bound_step4}
    \end{equation}
    
    Computing the conditional expectation $\mathbb{E}[(\cdot) | \mathcal{F}_t]$ for both sides we obtain
    \begin{equation}
        \begin{split}
            \mathbb{E}\left[\|\nabla G(\mathbf{y}_t) - \mathbf{d}_t\|^2 | \mathcal{F}_t \right] &= \rho_t^2 \mathbb{E}\left[\|\nabla G(\mathbf{y}_t) - \nabla \widehat{G_{z_t}^L (\mathbf{y}_t)}\|^2 | \mathcal{F}_t \right] + (1 - \rho_t)^2 \|\nabla G(\mathbf{y}_t) - \nabla G(\mathbf{y}_{t-1})\|^2 \\
            &\quad + (1 - \rho_t)^2 \|\nabla G(\mathbf{y}_{t-1})- \mathbf{d}_{t-1} \|^2 \\
            &\quad + 2\rho_t(1 - \rho_t)\mathbb{E}\Big[(\nabla G(\mathbf{y}_t) - \nabla \widehat{G_{z_t}^L (\mathbf{y}_t)})^T (\nabla G(\mathbf{y}_t) - \nabla G(\mathbf{y}_{t-1}))\Big] \\
            &\quad + 2\rho_t(1 - \rho_t)\mathbb{E}\Big[(\nabla G(\mathbf{y}_t) - \nabla \widehat{G_{z_t}^L (\mathbf{y}_t)})^T (\nabla G(\mathbf{y}_{t-1})- \mathbf{d}_{t-1})\Big]\\
            &\quad + 2(1 - \rho_t)^2(\nabla G(\mathbf{y}_t) - \nabla G(\mathbf{y}_{t-1}))^T (\nabla G(\mathbf{y}_{t-1})- \mathbf{d}_{t-1}).
        \end{split}
    \end{equation}
    
    Let's focus on the $\mathbb{E}\left[(\nabla G(\mathbf{y}_t) - \nabla \widehat{G_{z_t}^L (\mathbf{y}_t)})^T \allowbreak (\nabla G(\mathbf{y}_t) - \nabla G(\mathbf{y}_{t-1}))\right]$ term before moving further and call it $A$. By adding and subtracting $\nabla G_{z_t}(\mathbf{y}_t)$ inside $(\nabla G(\mathbf{y}_t) - \nabla \widehat{G_{z_t}^L (\mathbf{y}_t)})$ we obtain
    \begin{equation}
        \begin{split}
            A &= \mathbb{E}\Big[[(\nabla G(\mathbf{y}_t) - \nabla G_{z_t}(\mathbf{y}_t)) + (\nabla G_{z_t}(\mathbf{y}_t)) - \nabla \widehat{G_{z_t}^L (\mathbf{y}_t)}]^T  (\nabla G(\mathbf{y}_t) - \nabla G(\mathbf{y}_{t-1}))\Big]\\
            &= \mathbb{E}\Big[(\nabla G(\mathbf{y}_t) - \nabla G_{z_t}(\mathbf{y}_t))^T(\nabla G(\mathbf{y}_t) - \nabla G(\mathbf{y}_{t-1})) \\
            &+ (\nabla G_{z_t}(\mathbf{y}_t) - \nabla \widehat{G_{z_t}^L (\mathbf{y}_t)})^T (\nabla G(\mathbf{y}_t) - \nabla G(\mathbf{y}_{t-1}))\Big]\\
            &= \mathbb{E}\Big[(\nabla G(\mathbf{y}_t) - \nabla G_{z_t}(\mathbf{y}_t))^T(\nabla G(\mathbf{y}_t) - \nabla G(\mathbf{y}_{t-1}))\Big] \\
            &+ \mathbb{E}\Big[(G_{z_t}(\mathbf{y}_t) - \nabla \widehat{G_{z_t}^L (\mathbf{y}_t)})^T (\nabla G(\mathbf{y}_t) - \nabla G(\mathbf{y}_{t-1}))\Big].
        \end{split}
    \end{equation}
    Using the fact that $\nabla G_{z_t}(\mathbf{y}_t)$ is an unbiased estimator of the gradient $\nabla G(\mathbf{y}_t)$, i.e., $\mathbb{E}\left[\nabla G_{z_t}(\mathbf{y}_t) | \mathcal{F}_t \right] = \nabla G(\mathbf{y}_t)$, and replacing $(G_{z_t}(\mathbf{y}_t) - \nabla \widehat{G_{z_t}^L (\mathbf{y}_t)})$ with its upper bound $2\sqrt{n}\varepsilon (L)$ to obtain
    \begin{equation}
        A \leq \mathbb{E}\left[(2\sqrt{n}\varepsilon (L))^T (\nabla G(\mathbf{y}_t) - \nabla G(\mathbf{y}_{t-1}))\right].
    \end{equation}
    Applying a similar process to the $\mathbb{E}\left[(\nabla G(\mathbf{y}_t) - \nabla \widehat{G_{z_t}^L (\mathbf{y}_t)})^T (\nabla G(\mathbf{y}_{t-1})- \mathbf{d}_{t-1})| \mathcal{F}_t \right]$ term and using Young's inequality $\left(\langle \mathbf{a}, \mathbf{b}\rangle \leq \frac{1}{\beta}\frac{\|\mathbf{a}\|^2}{2} + \beta\frac{\|\mathbf{b}\|^2}{2}\right)$, also known as Peter-Paul inequality, to substitute the inner products with summations to obtain
    \begin{equation}
        \begin{split}
            \mathbb{E}\left[\|\nabla G(\mathbf{y}_t) - \mathbf{d}_t\|^2 | \mathcal{F}_t \right] &\leq \rho_t^2 \mathbb{E}\left[\|\nabla G(\mathbf{y}_t) - \nabla \widehat{G_{z_t}^L (\mathbf{y}_t)}\|^2 | \mathcal{F}_t \right] + (1 - \rho_t)^2 \|\nabla G(\mathbf{y}_t) - \nabla G(\mathbf{y}_{t-1})\|^2 \\
            &+ (1 - \rho_t)^2 \|\nabla G(\mathbf{y}_{t-1})- \mathbf{d}_{t-1} \|^2 \\
            &+ (\rho_t - \rho_t^2)\mathbb{E}\left[\beta_1 2\sqrt{n}\varepsilon (L) +  \frac{\|\nabla G(\mathbf{y}_t) - \nabla G(\mathbf{y}_{t-1})\|^2}{\beta_1}\right] \\
            &+ (\rho_t - \rho_t^2)\mathbb{E}\left[\beta_2 2\sqrt{n}\varepsilon (L) +  \frac{\|\nabla G(\mathbf{y}_{t-1})- \mathbf{d}_{t-1}\|^2}{\beta_2}\right]\\
            &+ (1 - \rho_t)^2\bigg(\beta_3\|\nabla G(\mathbf{y}_t) - \nabla G(\mathbf{y}_{t-1})^T\|^2 + \frac{1}{\beta_3}\|\nabla G(\mathbf{y}_{t-1})- \mathbf{d}_{t-1}\|^2\bigg),\\
            &= \rho_t^2 \mathbb{E}\left[\|\nabla G(\mathbf{y}_t) - \nabla \widehat{G_{z_t}^L (\mathbf{y}_t)}\|^2 | \mathcal{F}_t \right] + (1 - \rho_t)^2 \|\nabla G(\mathbf{y}_t) - \nabla G(\mathbf{y}_{t-1})\|^2 \\
            &+ (1 - \rho_t)^2 \|\nabla G(\mathbf{y}_{t-1})- \mathbf{d}_{t-1} \|^2 \\
            &+ (\rho_t - \rho_t^2)\left(\beta_1 2\sqrt{n}\varepsilon (L) +  \frac{\|\nabla G(\mathbf{y}_t) - \nabla G(\mathbf{y}_{t-1})\|^2}{\beta_1}\right) \\
            &+ (\rho_t - \rho_t^2)\left(\beta_2 2\sqrt{n}\varepsilon (L) +  \frac{\|\nabla G(\mathbf{y}_{t-1})- \mathbf{d}_{t-1}\|^2}{\beta_2}\right)\\
            &+ (1 - \rho_t)^2\bigg(\beta_3\|\nabla G(\mathbf{y}_t) - \nabla G(\mathbf{y}_{t-1})^T\|^2 + \frac{1}{\beta_3}\|\nabla G(\mathbf{y}_{t-1})- \mathbf{d}_{t-1}\|^2\bigg),\\
            &= \rho_t^2 \mathbb{E}\left[\|\nabla G(\mathbf{y}_t) - \nabla \widehat{G_{z_t}^L (\mathbf{y}_t)}\|^2 | \mathcal{F}_t \right] + \rho_t (1-\rho_t)(\beta_1 + \beta_2) 2\sqrt{n}\varepsilon (L) \\
            &+ \left((1-\rho_t)^2(1+\beta_3) + \frac{\rho_t(1-\rho_t)}{\beta_1}\right) \|\nabla G(\mathbf{y}_t) - \nabla G(\mathbf{y}_{t-1})\|^2 \\
            &+ \left((1 - \rho_t)^2(1+\frac{1}{\beta_3}) + \frac{\rho_t (1 - \rho_t)}{\beta_2}\right) \|\nabla G(\mathbf{y}_{t-1})- \mathbf{d}_{t-1} \|^2. 
        \end{split}
    \end{equation}
Since we assume that $\rho_t \leq 1$ we can replace all the $(1-\rho_t)^2$ terms by $(1-\rho_t)$. Applying this substitution and setting $\beta_1 =\rho_t$, $\beta_2 = 4$ and $\beta_3 = 4/\rho_t$, we get
    \begin{equation}
        \begin{split}
            \mathbb{E}\left[\|\nabla G(\mathbf{y}_t) - \mathbf{d}_t\|^2 | \mathcal{F}_t \right] &\leq \rho_t^2 \mathbb{E}\left[\|\nabla G(\mathbf{y}_t) - \nabla \widehat{G_{z_t}^L (\mathbf{y}_t)}\|^2 | \mathcal{F}_t \right] + \rho_t (1-\rho_t)(\rho_t + 4) 2\sqrt{n}\varepsilon (L)\\
            &\quad + 2(1-\rho_t)\left(1 + \frac{2}{\rho_t}\right) \|\nabla G(\mathbf{y}_t) - \nabla G(\mathbf{y}_{t-1})\|^2 \\
            &\quad + (1-\rho_t)\left(1+\frac{\rho_t}{2}\right) \|\nabla G(\mathbf{y}_{t-1})- \mathbf{d}_{t-1} \|^2.
        \end{split}
    \end{equation}
    Now using the inequalities $2(1-\rho_t)(1+(2/\rho_t)) \leq (4/\rho_t)$ and $(1-\rho_t)(1+(\rho_t/2)) \leq (1-(\rho_t/2))$ we obtain
    \begin{equation}
        \begin{split}
            \mathbb{E}\left[\|\nabla G(\mathbf{y}_t) - \mathbf{d}_t\|^2 \right] &\leq \rho_t^2 \mathbb{E}\left[\|\nabla G(\mathbf{y}_t) - \nabla \widehat{G_{z_t}^L (\mathbf{y}_t)}\|^2 \right] + \rho_t (1-\rho_t)(\rho_t + 4) 2\sqrt{n}\varepsilon (L)\\
            &+ \frac{4}{\rho_t} \|\nabla G(\mathbf{y}_t) - \nabla G(\mathbf{y}_{t-1})\|^2 + \left(1-\frac{\rho_t}{2}\right) \|\nabla G(\mathbf{y}_{t-1})- \mathbf{d}_{t-1} \|^2.
        \end{split}
    \end{equation}
 \sloppy   Now we need two auxiliary lemmas (Lemma~\ref{lem:pointwiseLipschitz} \& Lemma~\ref{lem:bias_bound}) to provide bounds for $\mathbb{E}\left[\|\nabla G(\mathbf{y}_t) - \nabla \widehat{G_{z_t}^L (\mathbf{y}_t)}\|^2 \right]$ and $\|\nabla G(\mathbf{y}_t) - \nabla G(\mathbf{y}_{t-1})\|^2$, respectively.
    
 \fussy   The term $\mathbb{E}\left[\|\nabla G(\mathbf{y}_t) - \nabla \widehat{G_{z_t}^L (\mathbf{y}_t)}\|^2 \right]$ can be bounded above by $(1 + \beta_0)\sigma_0^2 + \left(1 + \frac{1}{\beta_0}\right)2\sqrt{n}\varepsilon (L)$ according to Lemma~\ref{lem:bias_bound}. Based on Assumption~\ref{asm:boundedNorm} 
    and Lemma~\ref{lem:pointwiseLipschitz}, we can also show that the squared norm $\|\nabla G(\mathbf{y}_t) - \nabla G(\mathbf{y}_{t-1})\|^2$ is upper bounded by $f_{\max}^2rD^2/T^2$. Applying these substitutions yields
    \begin{equation*}
        \begin{split}
            \mathbb{E}\left[\|\nabla G(\mathbf{y}_t) - \mathbf{d}_t\|^2\right] &\leq \rho_t^2 (1 + \beta_0)\sigma_0^2 + \left(1 + \frac{1}{\beta_0} + \rho_t (1-\rho_t)(\rho_t + 4)\right)2\sqrt{n}\varepsilon (L) \\
            &\quad+ \frac{4}{\rho_t} f_{\max}^2rD^2/T^2 + \left(1-\frac{\rho_t}{2}\right)  \|\nabla G(\mathbf{y}_{t-1})- \mathbf{d}_{t-1} \|^2.
        \end{split}
    \end{equation*}
        
    We introduce one more Lemma~\ref{lem:recursive} to bound $\mathbb{E}\left[\|\nabla G(\mathbf{y}_t) - \mathbf{d}_t\|^2\right]$ recursively.
    
    Now define $\phi_t = \mathbb{E}\left[\|\nabla G(\mathbf{y}_t) - \mathbf{d}_t\|^2 | \mathcal{F}_t \right]$ and set $\rho_t = \frac{4}{(t+8)^{2/3}}$ to obtain
    \begin{equation*}
        \begin{split}
            \phi_t &\leq \left(1-\frac{2}{(t+8)^{2/3}}\right) \phi_{t-1} + \frac{16}{(t+8)^{4/3}} (1 + \beta_0)\sigma_0^2 \\
            &+ \Bigg(1 + \frac{1}{\beta_0} + \frac{4}{(t+8)^{2/3}} \left(1-\frac{4}{(t+8)^{2/3}}\right)\left(\frac{4}{(t+8)^{2/3}} + 4\right)\Bigg) 2\sqrt{n}\varepsilon (L) \\
            &+  \frac{f_{\max}^2rD^2(t+8)^{2/3}}{T^2}.
        \end{split}
    \end{equation*}
    Now use the conditions $8 \leq T$ and $t \leq T$ to replace $1/T$ by its upper bound $2/(t+8)$ and choose $\beta_0=1$:
    \begin{equation*}
        \begin{split}
            \phi_t \leq &\left(1-\frac{2}{(t+8)^{2/3}}\right) \phi_{t-1} + \frac{32}{(t+8)^{4/3}} \sigma_0^2 \\
            &+ \bigg(2 + \frac{4}{(t+8)^{2/3}} \left(1-\frac{4}{(t+8)^{2/3}}\right)\left(\frac{4}{(t+8)^{2/3}} + 4\right)\bigg) 2\sqrt{n}\varepsilon (L) +  \frac{4f_{\max}^2rD^2}{(t+8)^{4/3}}.
        \end{split}
    \end{equation*}
    Now using the inequality $\left(2 + \frac{4}{(t+8)^{2/3}} \left(1-\frac{4}{(t+8)^{2/3}}\right)\left(\frac{4}{(t+8)^{2/3}} + 4\right)\right) \leq \frac{112}{(t+8)^{2/3}}$ for $t \geq 0$
    \begin{equation*}
        \begin{split}
            \phi_t \leq &\left(1-\frac{2}{(t+8)^{2/3}}\right) \phi_{t-1} + \frac{32\sigma_0^2 + 224\sqrt{n}\varepsilon (L) + 4f_{\max}^2rD^2}{(t+8)^{4/3}}.
        \end{split}
    \end{equation*}
    Now using the result in Lemma~\ref{lem:recursive}, we obtain that 
    \begin{equation}
        \phi_t \leq \frac{Q}{(t+9)^{2/3}},
    \end{equation}
    where $Q = \max\{5\|\nabla G(\mathbf{y}_0 - \mathbf{d}_0)\|^2, 32\sigma_0^2 + 224\sqrt{n}\varepsilon (L) + 4f_{\max}^2rD^2\}.$ 
\end{proof}

Substitute $\mathbb{E}\left[\|\nabla G(\mathbf{y}_t) - \mathbf{d}_t\|^2\right]$ by its upper bound $Q/((t+9)^{2/3})$ according to the result of Lemma~\ref{lem:bounded_dt}. Further, set $\beta_t = \frac{Q^{1/2}}{2D(t+9)^{1/3}}$ and regroup the resulted expression to obtain
    \begin{equation}
        \begin{split}
            \mathbb{E}\left[G(\mathbf{y}^*) - G(\mathbf{y}_{t+1})\right] \leq &\left(1-\frac{1}{T}\right) \mathbb{E}\left[G(\mathbf{y}^*) - G(\mathbf{y}_t)\right] + \frac{2DQ^{1/2}}{(t+9)^{1/3}T} + \frac{f_{\max}r D^2}{2T^2}.\label{eq:maintheostep6}
        \end{split}
    \end{equation}
    By applying the inequality in (\ref{eq:maintheostep6}) recursively for $t = 0, \ldots, {T-1}$ we obtain
    \begin{equation}
        \begin{split}
            \mathbb{E}\left[G(\mathbf{y}^*) - G(\mathbf{y}_{T})\right] \leq &\left(1-\frac{1}{T}\right)^T \mathbb{E}\left[G(\mathbf{y}^*) - G(\mathbf{y}_0)\right] + \sum_{t=0}^{T-1} \frac{2DQ^{1/2}}{(t+9)^{1/3}T} + \sum_{t=0}^{T-1} \frac{f_{\max}r D^2}{2T^2}.\label{eq:maintheostep7}
        \end{split}
    \end{equation}
    
    Note that we can write
    \begin{equation}\label{eq:simple_sum}
        \begin{split}
            \sum_{t=0}^{T-1} \frac{1}{(t+9)^{1/3}} &\leq \frac{1}{9^{1/3}} + \int_{t=0}^{T-1} \frac{1}{(t+9)^{1/3}} dt\\
            &= \frac{1}{9^{1/3}} + \frac{3}{2} (t+9)^{2/3} \Big|_{t=T-1} - \frac{3}{2} (t+9)^{2/3} \Big|_{t=0}\\
            &\leq \frac{3}{2} (T+8)^{2/3} \leq \frac{15}{2} T^{2/3}
        \end{split}
    \end{equation}
    where the last inequality holds since $(T+8)^{2/3} \leq 5T^{2/3}$ for any $T \geq 1.$ By simplifying the terms on the right hand side of (\ref{eq:maintheostep7}) and using the inequality in (\ref{eq:simple_sum}) we can write
    \begin{equation}
        \begin{split}
            \mathbb{E}\left[G(\mathbf{y}^*) - G(\mathbf{y}_{T})\right] \leq &\frac{1}{e} \mathbb{E}\left[G(\mathbf{y}^*) - G(\mathbf{y}_0)\right] + \frac{15DQ^{1/2}}{T^{1/3}} + \frac{f_{\max}r D^2}{2T}.\label{eq:maintheostep8}
        \end{split}
    \end{equation}
    Here, we use the fact that $G(\mathbf{y}_0) \geq 0$, and hence the expression in (\ref{eq:maintheostep8}) can be simplified to 
    \begin{equation*}
        \mathbb{E}\left[G(\mathbf{y}_{T})\right] \geq (1-1/e) \mathbb{E}\left[G(\mathbf{y}^*))\right] - \frac{15DQ^{1/2}}{T^{1/3}} - \frac{f_{\max}r D^2}{2T},\label{eq:maintheostep9}
    \end{equation*}
    where $Q = \max\{5\|\nabla G(\mathbf{y}_0 - \mathbf{d}_0)\|^2, 32\sigma_0^2 + 224\sqrt{n}\varepsilon (L) + 4f_{\max}^2rD^2\}$ and $K = Q^{1/2} = \max\{3\|\nabla G(\mathbf{y}_0 - \mathbf{d}_0)\|^2, \sqrt{16\sigma_0^2 + 224\sqrt{n}\varepsilon (L)} + 2\sqrt{r}f_{\max}D\}.$
\end{proof}

\subsection{Custom Biases for the Problems in Sec.~\ref{sec: examples}}\label{app:customBias}
\subsubsection{Estimator Bias for Summarization Problems}
\label{app: proof_bias_bound}
\begin{theorem} \label{thm: epsilon_bound}
Assume a diversity reward function $f:~\{0,1\}^n \rightarrow \reals_+$ with $h(s)=\log(1+s)$. Then, consider the estimator $\widehat{\nabla G_z^L}(\mathbf{y}_K)$ given in (\ref{eq: poly_estimator}) using $\hat{f}_z^{L}(\vc{x})$,  the $L^{th}$ Taylor polynomial of $f(\vc{x})$ around $1/2$, given by \eqref{eq: taylor_f_iL}. Then, the bias of the estimator satisfies 
   $\big\|\nabla G_z(\vc{y}) - \widehat{\nabla G_z^L}(\vc{y})\big\|_2 \leq \frac{\sqrt{n}}{(L+1) 2^{L}}.$
\end{theorem}
\begin{proof}
    We begin by characterizing the residual error of the Taylor series of $h(s)=\log(1+s)$ around $1/2$:
    \begin{lemma} \label{lem: R_iL_bound}
    Let $\hat{h}^{L}(s)$ be the $L^{\text{th}}$ order Taylor approximation of $h(s) = \log(1 + s)$ around $1/2$, given by \eqref{eq: taylor_f_iL}. Then, $\hat{f}_z^L(\mathbf{x}) = \hat{h}^L(g_z(\mathbf{x}))$, satisfies Asm.~\ref{asmp:boundedPoly}, with:  
    \begin{equation}
        \varepsilon_{z}(L) =\frac{1}{(L+1) 2^{L+1}}.
    \end{equation}
    \end{lemma}
    \begin{proof}
    By the Lagrange remainder theorem,
    \begin{equation*}
    \begin{split}
     \left\lvert h(s)-\hat{h}^{ L}(s)\right\rvert&=\left\lvert\frac{h^{(L+1)}(s')}{(L+1)!} \left(s-\frac{1}{2}\right)^{L+1}\right\rvert \\
     &= \left\rvert \frac{\left(s-{1}/{2}\right)^{L+1}}{(L+1)\left(1+s'\right)^{L+1}} \right\lvert
    \end{split}
    \end{equation*}
    for some $s'$ between $s$ and $1/2$. Since $s \in [0, 1]$, (a) $|s-\frac{1}{2}|\leq \frac{1}{2}$, and (b) $s'\in[0, 1]$.  Hence
     $\left\lvert h(s) - \hat{h}^{L}(s) \right\rvert \leq \frac{1}{(L+1)2^{L+1}} .$
    \end{proof}
    To conclude the theorem, observe that:
    \begin{align*}
        \big\|\nabla G_z(\vc{y}) - \widehat{\nabla G_z^L}(\vc{y})\big\|_2 \leq 2\sqrt{n}\varepsilon (L) = \frac{\sqrt{n}}{(L+1) 2^{L}}.
    \end{align*}
\end{proof}

\subsubsection{Estimator Bias for Influence Maximization Problems}
\label{app: proof_bias_bound_IM}

\begin{theorem} \label{thm: epsilon_bound_IM}
For function $f:~\{0,1\}^n \rightarrow \reals_+$ that  given by $f(\mathbf{x}) = \mathbb{E}_{z \sim P} [f_z(\mathbf{x})]$, where  
consider the estimator $\widehat{\nabla G_z^L}$   given in (\ref{eq: poly_estimator}) using  $\hat{f}_z^L$, the $L^{\text{th}}$-order Taylor approximation of $f_z$ around $1/2$, given by \eqref{eq: taylor_f_iL}. Then, the bias of estimator $\widehat{\nabla G_z^L}$  satisfies
 $    \big\|\nabla G_z(\vc{y}) - \widehat{\nabla G_z^L}(\vc{y})\big\|_2 \leq \frac{\sqrt{n}}{(L+1) 2^{L}}.$ 
\end{theorem}
\begin{proof}
    To prove the theorem, observe that for all $\vc{y}\in [0,1]^n$:
\begin{equation*}
\begin{split}
\big\|\nabla G_z(\vc{y}) - \widehat{\nabla G_z^L}(\vc{y})\big\|_2 \leq 2\sqrt{n}\varepsilon (L) = \frac{\sqrt{n}}{(L+1) 2^{L}}.
\end{split}
\end{equation*}
\end{proof}

\subsubsection{Cache Networks (CN)\cite{mahdian2020kelly}}
\label{app:proof_bound_CN}
A Kelly cache network can be represented by a graph $G(V, E)$, $|E|=M$, service rates $\mu_j$, $j \in E$, storage capacities $c_v$, $v \in V$, a set of requests $\mathcal{R}$, and arrival rates $\lambda_r$, for $r \in \mathcal{R}$. Each request is characterized by an item $i^r\in\mathcal{C}$ requested, and a path $p^r\subset V$ that the request follows.  For a detailed description of these variables, please refer to \cite{mahdian2020kelly}. Requests are forwarded on a path until they meet a cache storing the requested item. In steady-state,  the traffic load on an edge $(u,v)$ is given by  
\begin{equation} \label{eq:CNgi}
    g_{(u, v)}(\vc{x}) = \frac{1}{\mu_{u, v}}\sum_{r \in \mathcal{R}:(v, u)\in p^r} \lambda^r \prod_{k'=1}^{k_{p^r}(v)}(1-x_{p_k^r, i^r}).
\end{equation}
where $\vc{x}\in \{0,1\}^{|V||\mathcal{C}|}$ is a vector of binary coordinates $x_{vi}$ indicating if $i\in \mathcal{C}$ is stored in node $v\in V$. If $s$ is the load on an edge,  the expected total number of packets in the system is given by $h(s)=\frac{s}{1-s}$.
Then using the notation $z=(u, v) \in E$ to index edges, the expected total number of packets in the system in steady state can indeed be written as $\mathbb{E}_{z \sim P}\left[h(g_z(\vc{x}))\right]$ \cite{mahdian2020kelly}. 
Mahdian et al.~maximize the  \emph{caching gain} $f: \{0, 1\}^{|V||\mathcal{C}|} \rightarrow \reals_+$ as 
\begin{equation} \label{eq:CN}
    f(\vc{x}) = \textstyle\mathbb{E}_{z \sim P} \left[h(g_z(\vc{0})) -  h(g_z(\vc{x}))\right]
\end{equation}
subject to the capacity constraints in each class.
The caching gain $f(\vc{x})$ is  monotone and submodular, and the capacity constraints form a partition matroid \cite{mahdian2020kelly}. 
Moreover, $h(s)=\frac{s}{1-s}$ can be approximated within arbitrary accuracy by its $L^{\text{th}}$-order Taylor approximation around $0$, given by:
\begin{equation} \label{eq: f_iL_CN}
    \hat{h}^{L}(s) = \textstyle\sum_{\ell = 1}^L s^\ell
\end{equation}
We show in 
App.~\ref{app:proof_bound_CN} 
that this estimator ensures that $f$ indeed satisfies Asm.~\ref{asmp:boundedPoly}. Proof of this lemma can be found in App.~\ref{app:proof_bound_CN}.
Furthermore, we  bound the estimator bias appearing in Thm.~\ref{thm:main} as follows:
\begin{theorem} \label{thm:epsilon_bound_CN}
Assume a caching gain function $f:~\{0,1\}^{|V||\mathcal{C}|} \rightarrow \reals_+$ that is given by (\ref{eq:CN}). Then, consider Algorithm \ref{alg: SCG} in which $\nabla G(\mathbf{y}_K)$ is estimated via the polynomial estimator given in (\ref{eq: poly_estimator}) where $\hat{f}_z^{L}(\vc{x})$ is the $L^{th}$ Taylor polynomial of $f(\vc{x})$ around $0$. Then, the bias of the estimator is bounded by
\begin{equation}
    \big\|\nabla G_z(\vc{y}) - \widehat{\nabla G_z^L}(\vc{y})\big\|_2 \leq 2\sqrt{{|V||\mathcal{C}|}}\frac{\bar{s}^{L+1}}{1-\bar{s}},
\end{equation}
where $\bar{s}<1$ is the largest load among all edges when caches are empty.  
\end{theorem}
\begin{proof}
\begin{lemma} \label{lem:bound_CN}
Let $\hat{h}^{L}(s)$ be the $L^{th}$ Taylor polynomial of $h(s)=\frac{s}{1-s}$ around $0$. Then, $h(s)$ and its polynomial estimator of degree $L$, $\hat{h}^{L}(s)$, satisfy Asm.~\ref{asmp:boundedPoly} where 
\begin{equation}
    \varepsilon(L) = \frac{\bar{s}^{L+1}}{1 - \bar{s}}.
\end{equation}
\end{lemma}
\begin{proof}
$L^{th}$ Taylor polynomial of $h(s)$ around $0$ is
\begin{equation}
    \hat{h}_{L}(s) = \textstyle\sum_{l = 0}^L \frac{h^{(\ell)}(0)}{\ell!} s^{\ell} = \sum_{\ell=1}^{L} s^{\ell}
\end{equation}
where $h^{(\ell)}(s) = \frac{\ell!}{(1-s)^{\ell+1}}$ for $h(s) = \frac{s}{1-s}$. 
\begin{align*}
    h(s) &= \frac{s}{1 - s} = \textstyle\sum_{\ell=1}^{\infty} s^{\ell} = \sum_{\ell=1}^{L} s^{\ell} + \sum_{\ell=L+1}^{\infty} s^{\ell} \\
    &= \textstyle\sum_{\ell=1}^{L}s^{\ell}+s^L\sum_{\ell=1}^{\infty}s^{\ell}=\sum_{\ell=1}^{L} s^{\ell} + \frac{s^{L+1}}{1-s}
\end{align*}
Then, the bias of the Taylor Series Estimation around $0$ becomes:
 \begin{align*}
    \left| \frac{s}{1 - s} - \textstyle\sum_{\ell=1}^{L} s^{\ell}\right|  =  \frac{s^{L+1}}{1-s} \leq \frac{\bar{s}^{L+1}}{1 - \bar{s}} = \varepsilon(L).
\end{align*}
for all $s \in [0, \bar{s}]$ where $\bar{s} = \max_{z \sim P} s_z$. \hspace{\stretch{1}} \qed


Since $\lim_{L\to \infty} \frac{\bar{s}^{L+1}}{1 - \bar{s}} = 0$, for all $\bar{s} \in [0, 1)$, Taylor approximation gives an approximation guarantee for maximizing the queue size function by Asm.~\ref{asmp:boundedPoly},  where the error of the approximation is given by Lem.~\ref{lem:gradientBias} as
\begin{align*}
\big\|\nabla G_z(\vc{y}) - \widehat{\nabla G_z^L}(\vc{y})\big\|_2 \leq 2\sqrt{|V||\mathcal{C}|}\varepsilon (L) = 2\sqrt{|V||\mathcal{C}|}\frac{s^{L+1}}{1-s}.
\end{align*} Then, $\varepsilon(L) \leq 2\sqrt{{|V||\mathcal{C}|}}\frac{\bar{s}^{L+1}}{1-\bar{s}}$.
\end{proof}
\end{proof}

\subsection{Experimentation Details}\label{app:exps}

\textbf{Synthetic Datasets.} We generate directed bipartite graph instances with number of instances $|z| = 1$, $5$, $10$, $100$ and number of nodes $n = 200$, $400$, $1000$. Nodes are equally divided into left ($V_1$) and bottom ($V_2$) nodes where $|V_1| = |V_2|$. The seeds are always selected from $V_1$ and edges are placed between $V_1$ and $V_2$ so that the degrees of the nodes either follow a uniform or power law distribution. 

\noindent\textbf{Real Datasets.} We use two real-world datasets \texttt{Zachary Karate Club} (\texttt{ZKC}) \cite{zachary1977information} and \texttt{MovieLens} \cite{movielens}.


\noindent\textbf{Influence Maximization. } We experiment on three synthetic datasets and one real dataset. For synthetic data, We generate directed bipartite graph instances with number of instances $|z| = 1$, $5$, $10$, $100$ and number of nodes $n = 200$, $400$, $1000$. Nodes are equally divided into left ($V_1$) and bottom ($V_2$) nodes where $|V_1| = |V_2|$. The seeds are always selected from $V_1$ and edges are placed between $V_1$ and $V_2$ so that the degrees of the nodes 
follow 
power law (\texttt{SyntheticBipartitePowerlaw}) distribution. We construct a partition matroid of $m=4$ equal-size partitions of $V_1$ and set $k=1$ or $2$ elements from each partition. The real dataset is the \texttt{Zachary Karate Club} (\texttt{ZKC}) \cite{zachary1977information}. We generate $|z| = 20$ cascades following the independent cascade model \cite{kempe2003maximizing} using \texttt{Network Diffusion Library} \cite{rossetti2017ndlib}. The probability for each node to influence its neighbors is set to $p=0.5$. We divide the dataset into two partitions following its existing labels and set $k=3$.

\noindent\textbf{Facility Location. } For \emph{facility location} problems, we experiment on the \texttt{MovieLens} dataset \cite{movielens}. it has 1M ratings from $n = 6041$ users for $|z| = 4000$ movies. We treat movies as facilities, users as customers and ratings as $w_{i, j}$. We divide the movies into $m = 10$ partitions based on the first genre name listed for each movie and finally we set $k=2$. 

\end{document}